\def\ARXIVVERSION{1}

\documentclass{article} %

\usepackage{iclr2027_conference,times}

\usepackage{etoolbox}

\usepackage{amsmath,amsfonts,bm}

\def\eqref#1{equation~\ref{#1}}

\def\1{\bm{1}}

\DeclareMathAlphabet{\mathsfit}{\encodingdefault}{\sfdefault}{m}{sl}
\SetMathAlphabet{\mathsfit}{bold}{\encodingdefault}{\sfdefault}{bx}{n}

\usepackage{hyperref}
\hypersetup{
  colorlinks = true,
  linkcolor  = blue,
  citecolor  = blue,
  urlcolor   = blue,
  filecolor  = blue
}
\usepackage[capitalize,noabbrev,nameinlink]{cleveref}
\usepackage{url}
\usepackage{graphicx}
\usepackage{subcaption}
\usepackage{booktabs}

\usepackage[most]{tcolorbox}
\usepackage{xcolor}
\usepackage{colortbl}

\definecolor{ourrow}{gray}{0.85}
\definecolor{findingbg}{RGB}{247,248,250}
\definecolor{findingborder}{RGB}{185,190,198}

\newtcolorbox{findingbox}{
  colback=findingbg,
  colframe=findingborder,
  boxrule=0.5pt,
  arc=1.5pt,
  left=6pt,
  right=6pt,
  top=5pt,
  bottom=5pt,
  before skip=9pt,
  after skip=9pt,
  boxsep=0pt
}

\usepackage{color-edits}
\addauthor{ab}{red}

\usepackage{listings}
\lstdefinestyle{prompt}{
    basicstyle=\fontsize{7.5}{9.0}\selectfont\rmfamily,,
    columns=fullflexible,
    keepspaces=false,
    showstringspaces=false,
    breaklines=true,
    breakatwhitespace=true,
    breakindent=0pt,
    breakautoindent=false,
    aboveskip=0pt,
    belowskip=0pt,
}

\newtcblisting{earlyprompt}[1]{
    enhanced,
    listing only,
    listing engine=listings,
    listing options={style=prompt},
    title={#1},
    fonttitle=\bfseries\footnotesize,
    coltitle=black,
    colback=black!1,
    colframe=black!18,
    colbacktitle=blue!5,
    boxrule=0.45pt,
    arc=1.2mm,
    left=1.2mm,
    right=1.2mm,
    top=1.0mm,
    bottom=1.0mm,
    before skip=5pt,
    after skip=5pt,
}

\newtcblisting{lateprompt}[1]{
    enhanced,
    listing only,
    listing engine=listings,
    listing options={style=prompt},
    title={#1},
    fonttitle=\bfseries\footnotesize,
    coltitle=black,
    colback=black!1,
    colframe=black!18,
    colbacktitle=orange!6,
    boxrule=0.45pt,
    arc=1.2mm,
    left=1.2mm,
    right=1.2mm,
    top=1.0mm,
    bottom=1.0mm,
    before skip=5pt,
    after skip=5pt,
}

\title{Rethinking LLM Evaluation Over Time}
\title{Why You Should View LLM Evaluation as a Forecasting Problems}
\title{LLM Evaluation As Forecasting}
\title{LLM Evaluation is a Forecasting Problem}

\author{
Rahul Thomas\textsuperscript{1} \qquad
Adam Block\textsuperscript{1,2}\thanks{%
\protect\begin{tabular}[t]{@{}l@{}}
Equal advising. \\
\llap{\textsuperscript{1}\,}Department of Electrical Engineering, Columbia University. \\
\llap{\textsuperscript{2}\,}Department of Computer Science, Columbia University. \\
Correspondence to {
{\fontsize{8}{9.5}\selectfont\ttfamily\bfseries
rkt2129@columbia.edu, abb2190@columbia.edu, mig2132@columbia.edu.}
}
\protect\end{tabular}%
} \qquad
Micah Goldblum\textsuperscript{1,2}\footnotemark[1]
}

\ifdefined\ARXIVVERSION
    \iclrfinalcopy
\fi

\begin{document}

\ifdefined\ARXIVVERSION
    \makeatletter
    \patchcmd{\@maketitle}{\LARGE\sc}{\LARGE\bfseries}{}{}
    
\patchcmd{\@maketitle}
  {\hsize\textwidth}
  {\hsize\textwidth\centering}
  {}{}
    \patchcmd{\section}{\large\sc\raggedright}{\large\bfseries\raggedright}{}{}
    \patchcmd{\subsection}{\normalsize\sc\raggedright}{\normalsize\bfseries\raggedright}{}{}
    \patchcmd{\abstract}
  {\large\sc Abstract}
  {\large\bfseries Abstract}
  {}{}

    \makeatother
\fi

\maketitle

\ifdefined\ARXIVVERSION
    \lhead{Preprint}
\fi

\begin{abstract}
Standard procedure in machine learning is to randomly split train and test data, emulating IID draws from one underlying data distribution. In practice, models are trained on past data and deployed in the future, which may come from distinct distributions. In this paper, we find that web text is indeed highly non-stationary, as text distributions shift over time. We further observe that model performance changes continuously over time across diverse benchmarks. Consequently, future model performance is better predicted by measurements made on recent evaluation data, and is best predicted by time series forecasters applied to historic performance rather than by historic performance itself. Choosing which model to deploy based on forecasted accuracy enables more reliable selection of performant models, thus improving downstream accuracy in the regime practitioners care about: deployment. Taken together, our experiments suggest that time should be treated as a fundamental variable in LLM evaluation. We release our \href{https://github.com/rthomas22c/LLM-Evaluation-is-a-Forecasting-Problem}{code} and \href{https://huggingface.co/datasets/rkthomas/llm-forecasting-datasets}{data}.

\end{abstract}

\section{Introduction}
\label{sec:introduction}

Among the first lessons a machine learning student learns is to randomly split their dataset into train and validation splits.  This convention is well-motivated by \emph{exchangeability}: if we assume that samples are independent draws from the same distribution, loss computed on a held-out validation set provides an unbiased estimate of expected loss on new samples during deployment as long as they are drawn from the same distribution \citep{shalev2014understanding}.

The importance of minimizing distribution shift between evaluation and deployment predates modern machine learning benchmarks. In an early NIST handwriting-recognition competition, the training data were written by Census workers, whereas the official test data were collected from high-school students; systems achieving less than 1\% error on validation sets drawn from the training distribution performed substantially worse on the official test set. \citet{lecun1995learning} subsequently repartitioned the NIST data when constructing MNIST, splitting examples from both source populations across training and test sets while keeping writers disjoint to emulate identically distributed samples.

Large language models (LLMs) violate the assumption that their training, evaluation, and deployment samples are drawn from the same distribution. 
LLM training data is gathered from historic web scrapes, and a large portion of the data is born years before training commences \citep{brown2020language,gao2020pile,lazaridou2021mind}, in contradistinction to data seen during deployment, which is often produced after the training cutoff. In this paper, we investigate the extent to which the temporal distribution shift affects how we should be evaluating LLMs and, by treating evaluation as a forecasting problem, propose best practices for mitigating this shift and improving estimates of model quality during deployment.

We begin by observing that text distributions are highly nonstationary under a wide variety of statistical metrics and in a manner that heavily influences LLM training and evaluation. The nonstationarity of text data distributions makes LLM evaluation closer to \emph{time series forecasting} than to classical i.i.d. estimation. Inspired by \emph{backtesting} from the financial engineering community \citep{white2000reality, bailey2017probability}, we treat performance as a time series and ask how well measurements available at one point in time predict performance on later data. Even though randomly shuffling data across dates erases precisely the dependence that is useful for predicting future performance, standard LLM benchmarks commonly collapse examples across many dates into one time-agnostic score, including old samples which poorly predict future performance and obscure temporal structure.

\begin{figure}[t!]
    \centering
    \includegraphics[width=\linewidth]{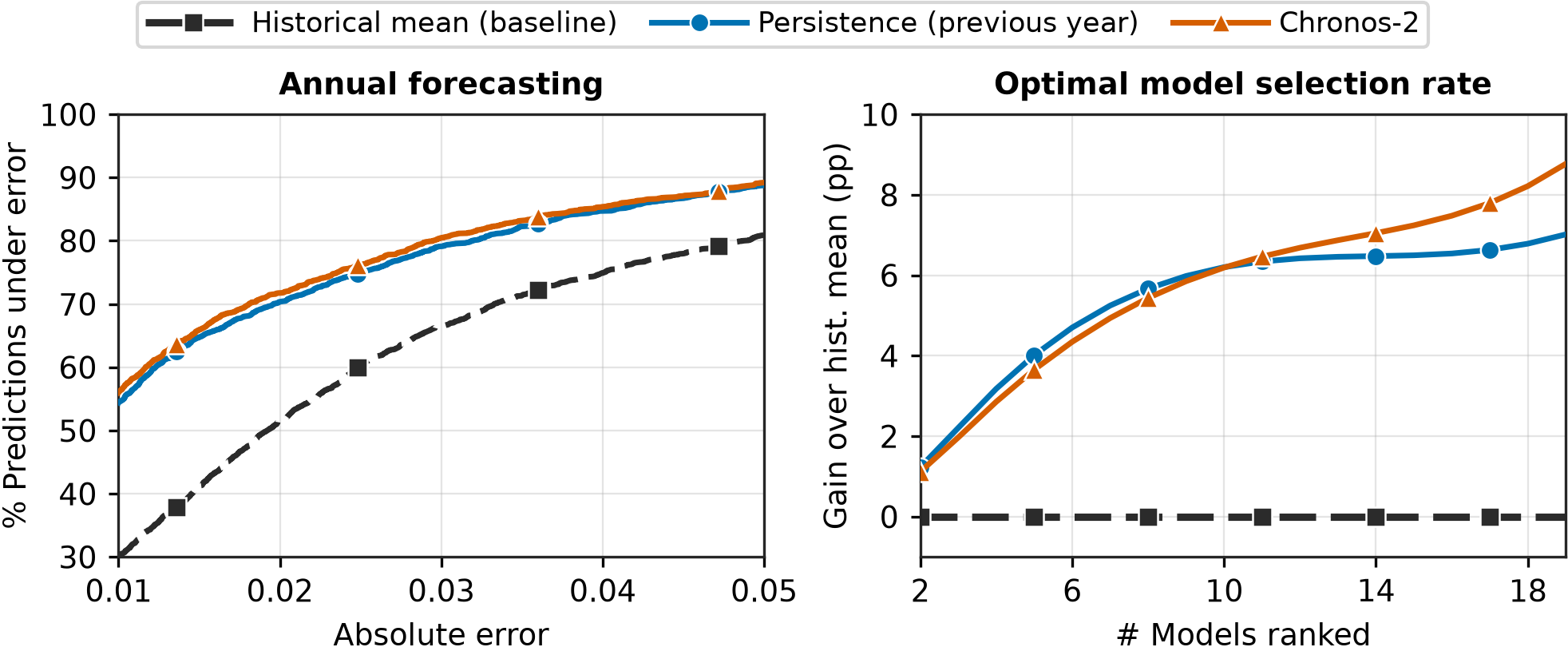}
    \caption{\textbf{Recent evaluations and time series forecasting outperform temporally pooled benchmark scores for both future performance prediction and model selection.} We evaluate 19 open-source models on 12 temporally labeled datasets, using the historical mean as the pooled baseline. \textit{(Left)} The most recent year and Chronos-2 yield fewer predictions exceeding each error threshold. \textit{(Right)} Both improve annual model selection, with larger gains as the candidate set grows.}
    \label{fig:main-figure}
\end{figure}

Instead, we propose applying forecasting algorithms, like Chronos-2 \citep{ansari2025chronos} or Simple Exponential Smoothing (SES) \citep{holt1957forecasting}, to the time series of historic model accuracy computed on data from different months or years in order to better predict future performance than historic model accuracy itself. \Cref{fig:main-figure} demonstrates the efficacy of this approach in reducing prediction error relative to standard practice.  While improved forecasting is interesting in its own right, one of its most important applications is \emph{model selection}: given a task and several models, which one should we deploy?  We find that the forecasting methods that leverage temporal information lead to substantially better deployment decisions than the temporally agnostic baseline, demonstrating the concrete benefits of treating evaluation as a forecasting problem.

We discuss related work in \cref{appendix:related-work}. Our contributions are summarized as follows:
\begin{itemize}
    \item \textbf{We observe that web text is strongly non-stationary over time.} Across three popular pretraining corpora, lexical, topical, syntactic, and semantic properties shift systematically, and models generalize best between temporally adjacent train and test data.

    \item \textbf{We show that evaluation staleness matters.} Across 19 open-source models and 12 benchmarks, recent evaluations consistently predict near-future model performance better than older ones, with stale evaluations producing larger tail errors.

    \item \textbf{We frame LLM evaluation as a forecasting problem.} Time-aware predictors, such as simple persistence and Chronos-2, outperform the standard practice of averaging all benchmark scores for predicting future model performance.

    \item \textbf{We show that better forecasts improve model selection for deployment.} Time-aware evaluation increases the rate of selecting the best future-performing model and reduces the frequency of high-regret model choices.
\end{itemize}

\section{Temporal Shifts in Pretraining Data}
\label{sec:temporal-shifts}

We first investigate temporal distributional shifts in pretraining datasets. In \cref{subsec:dataset-metrics}, we show that pretraining corpora derived from Common Crawl are \textbf{non-stationary} by analyzing lexical metrics, domain shift, and embedding drift. Then, in \cref{subsec:yearly-pretrain}, we show that model generalization degrades smoothly with temporal separation between train and test data. We further demonstrate that this degradation is \textbf{directional}, with a greater price in generalization paid when training on past data and evaluating on future data than in the reverse direction.

\subsection{Pretraining Datasets are Non-Stationary}
\label{subsec:dataset-metrics}

\begin{figure}[!t]
    \centering
    \includegraphics[width=\linewidth]{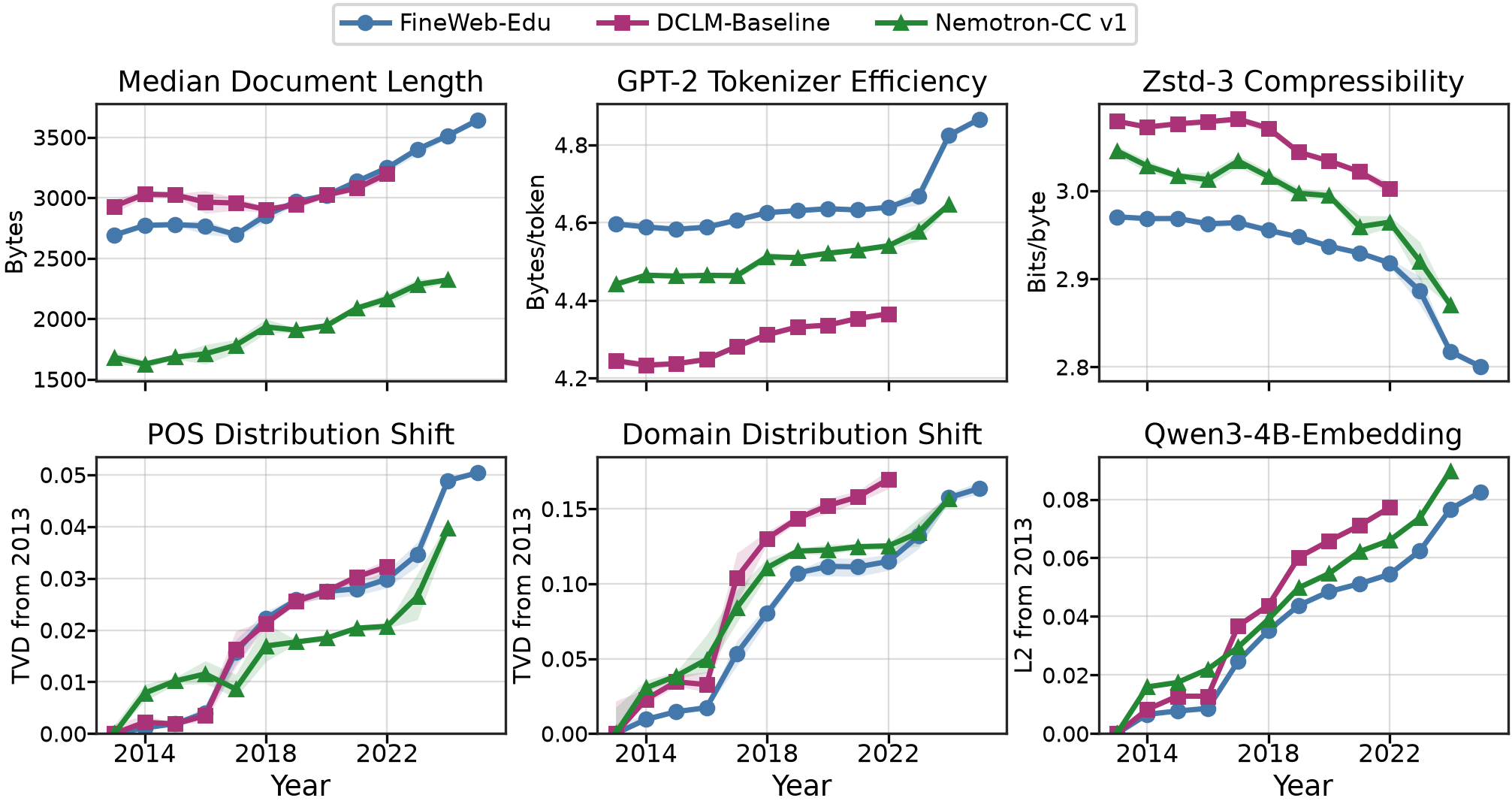}
    \caption{\textbf{Surface-level, syntactic, topical, and semantic properties of pretraining corpora systematically shift over time.} Across FineWeb-Edu, DCLM-Baseline, and Nemotron-CC v1, document length and tokenizer efficiency increase while byte compressed-size ratio decreases. Part-of-speech and domain distributions, along with annual document-embedding centroids, increasingly diverge from their initial values. Shading shows bootstrapped 95\% confidence intervals.}
    \label{fig:pretraining-trends}
\end{figure}

Pretraining corpora contain web data collected across many years, but these temporal slices are often treated as samples from a common underlying distribution. We first question this assumption: do the characteristics and composition of pretraining data remain stable over time?

We study the pretraining corpora FineWeb-Edu \citep{penedo2024fineweb}, DCLM-Baseline \citep{li2024datacomp}, and Nemotron-CC v1\footnote{We do not use Nemotron v2 as it has no date-bearing field. Further, we restrict Nemotron to the three tiers in its organic partition (high, medium-high, medium), and we do not include synthetic or translated data.} \citep{su2024nemotron}, which span Common Crawl data from 2013--2025, 2013--2022, and 2013--2024, respectively. We use these datasets because they have varying filtering pipelines, which allows us to test whether temporal trends persist across distinctly curated corpora.

We begin by examining length and compressibility, which are related to heuristics used for filtering pretraining data \citep{penedo2024fineweb,soldaini2024dolma}. %
Because years contain varying numbers of dumps with varying sizes, we aggregate by year. In \cref{fig:pretraining-trends}, we find that median document length and GPT-2 tokenizer efficiency increase over time and zstd-3 compressed-size ratio decreases over time for all datasets, while \cref{fig:pretraining-trends-more} shows the same trend for GPT-2 compressibility. Thus, while documents lengthen over time, %
their text becomes more compressible. We emphasize that the consistency of these trends across datasets suggests that they are not tied to a particular filtering pipeline.

Next, we ask whether these  shifts accompanied by changes in data composition. Such changes would indicate that models encounter different types of content as the temporal range of their pretraining data changes. We measure shifts in part-of-speech (POS) and domain distributions relative to the first year of each corpus in \cref{fig:pretraining-trends}. For domains, we adopt the fastText domain classifier from OLMo-3 \citep{olmo2025olmo}. We find that the distance of both distributions from the 2013 baseline increases nearly monotonically over time, and \cref{fig:pretraining-trends-more} shows the same trend in raw token distributions. Some domain shifts can be attributed to explicit events.  For example, the sharpest yearly changes for all corpora occur at 2017, where the distinct host count doubles for FineWeb and triples for DCLM and Nemotro, which coincides with an expansion of Common Crawl's official coverage: in 2017, Common Crawl added URLs within two hops of the Alexa top-million domains \citep{nagel2017january}. Thus, temporal shifts can arise from changes in crawl coverage, not just changes in the web.

Finally, we test whether drift also appears in semantic text embeddings. In \cref{fig:pretraining-trends}, we embed documents from each year with Qwen3-4B-Embedding \citep{zhang2025qwen3}, and we plot L2 distance between the annual centroid and the 2013 baseline centroid over time. In \cref{fig:pretraining-trends-more}, we also plot this drift with Qwen3-0.6-Embedding and MiniLM-L6-v2 \citep{sentenceTransformers2021allMiniLM}. Across all encoder-dataset pairs, distance increases nearly monotonically over time. Due to the difficulty of interpreting centroid shifts in absolute terms, we compare these shifts to \emph{inter-corpora} distances. We find that temporal drift within a corpus can rival distance between corpora, and it is largely driven by a shift in embedding location rather than local density (\cref{appendix:additional-metrics}).

In addition to investigating the ways in which corpus distributions shift across time, we also consider a possible source of these temporal shifts: sharp trends in later years are consistent with increased LLM-generated text. In \cref{fig:pretraining-trends-more}, we run the Desklib AI text detector \citep{desklib2025aitextdetector} and aggregate dump-level probabilities by year. For FineWeb and Nemotron, we find the estimated probability of AI-generated text is flat before the ChatGPT release at the end of November 2022 and rises sharply thereafter. The same trend appears in the frequency of four markers\footnote{These markers are ``delve", ``as a large language model", ``it's important to note", and ``rich tapestry", originally defined and used by the FineWeb blog \citep{penedo2024finewebblog} in their temporal analysis.} associated with LLM-generated text; we do not analyze DCLM as its coverage does not extend into the LLM era. Because our other temporal trends begin well before 2022, increased LLM-generated text cannot fully explain our observed long-term shifts, but may contribute to their acceleration in recent years.

We remark that these trends are not clearly explained by filtering mechanisms. FineWeb's mean educational score and DCLM's mean fastText score are nearly constant across years, while Nemotron's quality composition only materially changes in the last two years (\cref{fig:pretraining-trends-more}). Thus, changes in the datasets' filtering signals do not appear sufficient to explain the observed temporal shifts.

\begin{findingbox}
\textbf{Observation 1: Web pretraining data are non-stationary.}
We find that lexical, information-theoretic, syntactic,
topical, and semantic metrics applied to pretraining data shift 
over time.%
\end{findingbox}

\subsection{Generalization Across Time Degrades with Temporal Separation}
\label{subsec:yearly-pretrain}

While we have established substantial distributional shift in text data, it is not a priori clear that such shifts impact model performance. We now show that these dsitributional shifts can indeed negatively affect genearlization performance of language models.  For each studied dataset, we independently pretrain a 411M model on each year at the Chinchilla-optimal token budget following DCLM conventions \citep{li2024datacomp}, with precise details in \cref{appendix:training-details}. Let $L(s,t)$ denote the year $t$ test loss of a model pretrained on data from year $s$. Because absolute test loss can vary substantially across years, we report the \emph{excess loss} $\Delta L(s,t)=L(s,t)-L(t,t)$ relative to a model pretrained on the test year's distribution, which normalizes for test year difficulty and focuses on the impact of temporal train-test mismatch.

\begin{figure}[t!]
    \centering

    \includegraphics[width=\textwidth]{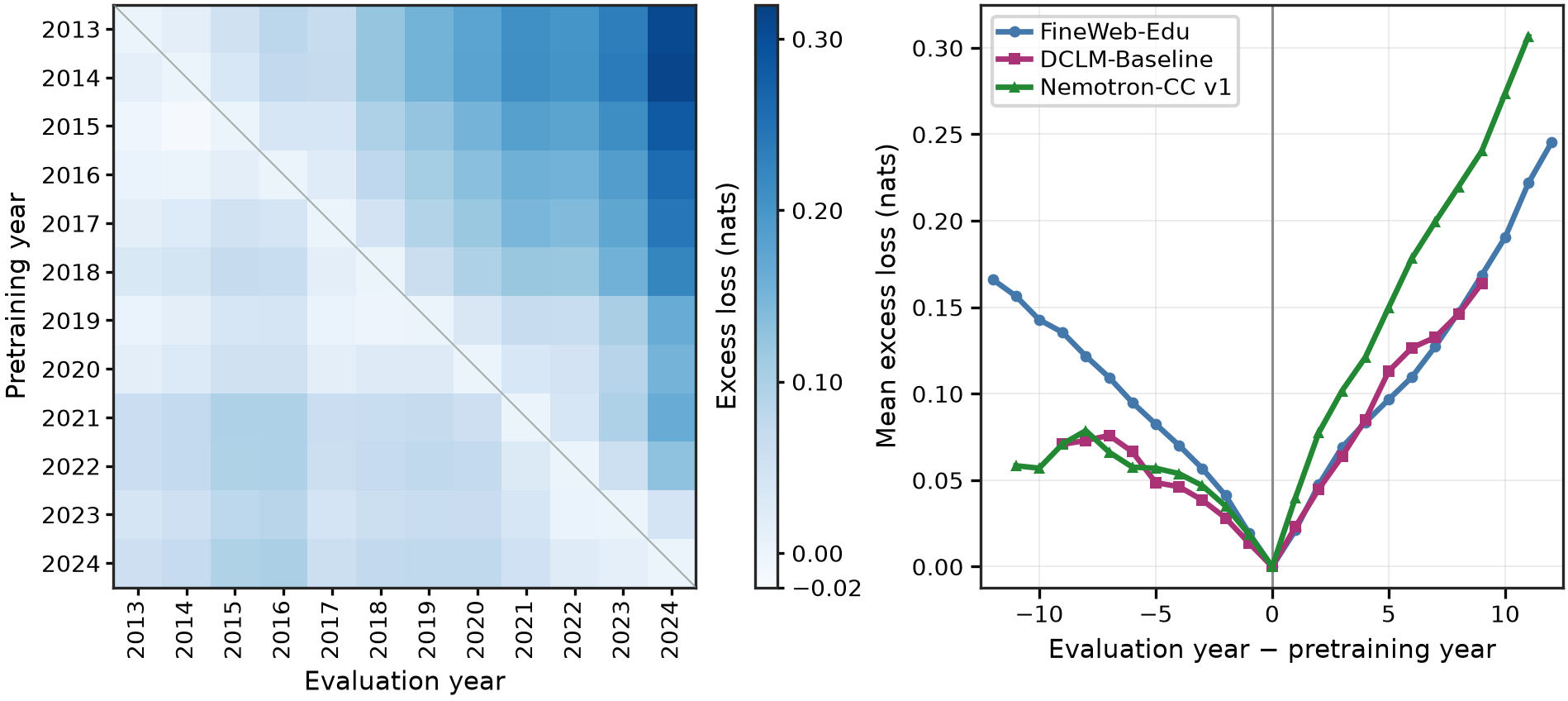}

    \caption{\textbf{Models generalize best between nearby years, with degradation increasing more rapidly when generalizing to future data.} \textit{(Left)} Excess loss $\Delta L(s,t)=L(s,t)-L(t,t)$ for Nemotron-CC v1 is smallest near the diagonal and increases with temporal separation; FineWeb-Edu and DCLM-Baseline are in \cref{appendix:yearly-heatmaps}. \textit{(Right)} Averaging by year offset $t-s$ shows the same pattern across all corpora, with larger excess loss when generalizing to future data (positive offsets).}
    \label{fig:model-transfer}
\end{figure}

Across all three corpora, excess loss $\Delta L(s,t)$ is smallest when pretraining and evaluation years are close, and grows smoothly with temporal separation $|t-s|$. \cref{fig:model-transfer} (left) shows this \textbf{temporal locality}: the Nemotron heatmap is lowest near the diagonal $t\approx s$, and aggregating model-evaluation pairs by signed year offset $t-s$ shows degradation in excess loss with increasing temporal separation. Thus, models generalize better between nearby years than between distant years. 

Temporal effects are also \textbf{asymmetric}, with greater degradation when the training year is earlier than the evaluation year, which makes sense in light of the unidirectional nature of time. This effect is starkly shown in \Cref{fig:model-transfer} (right), where, for each pair of years, we compare transfer from the earlier year to the later year against the reverse direction. This asymmetry matters in practice because deployment is directional: future performance must be estimated using only data from the past, which is empirically the more difficult estimation problem. Together, non-stationarity and temporal locality suggest that the predictive value of evaluation data should depend on its age. In particular, recent evaluations may better estimate near-future model performance than older evaluations.

\begin{findingbox}
\textbf{Observation 2: %
Models generalize better between nearby years than distant ones.} This degradation is asymmetric; transfer from past training to future eval data is worse than the reverse.
\end{findingbox}

\section{Time-Aware Evaluation Predicts Future Model Performance}
\label{sec:future-evals}

In \cref{sec:temporal-shifts}, we showed that pretraining corpora evolve over time and that models generalize better among nearby years than among more distant years. We now ask whether a similar temporal structure appears in model \emph{evaluations}, and whether accounting for it allows us to better predict future model performance. As benchmark scores are used to inform model comparisons and deployment decisions, if older evaluations become less predictive of future performance, then pooling them with recent observations can distort predicted performance and even reverse model rankings.

We study 12 datasets with temporal metadata, spanning biomedical QA, classification, summarization, citation prediction, and other scientific and professional domains (\cref{tab:datasets}). 
Full dataset details are in \cref{appendix:datasets}, and all models and inference settings are described in \cref{appendix:model-generation}.

\begin{table*}[t]
\centering
\caption{We evaluate frontier and open-source models on 12 datasets of varying scale, task type, and temporal coverage. We construct all shaded datasets from the relevant source corpus (\cref{appendix:datasets}).}
\label{tab:datasets}

\setlength{\tabcolsep}{5pt}
\renewcommand{\arraystretch}{0.96}

\begin{tabular}{llrrr}
\toprule
\textbf{Dataset} &
\textbf{Task Description} &
\textbf{Start Year} &
\textbf{End Year} &
\textbf{Examples} \\
\midrule
BioASQ            & Biomedical QA                    & 2014 & 2025 & 4,067 \\
FinQA             & Financial QA                     & 1999 & 2019 & 8,281 \\
StreamingQA       & Evolving knowledge QA            & 2007 & 2020 & 16,307 \\
\rowcolor{ourrow} SCIDOCA           & Scientific citation prediction   & 2012 & 2025 & 70,000 \\
\rowcolor{ourrow} NVD-VSP           & CVSS severity scoring             & 2019 & 2025 & 116,711 \\
\rowcolor{ourrow} NVD-RCM           & CWE root-cause classification     & 2017 & 2025 & 132,103 \\
\rowcolor{ourrow} CLERC             & Legal citation matching           & 2006 & 2019 & 140,000 \\
\rowcolor{ourrow} PubMed-MeSH       & MeSH descriptor prediction        & 2010 & 2025 & 160,000 \\
\rowcolor{ourrow} PubMed-Rhetorical & Rhetorical-role classification    & 2010 & 2025 & 160,000 \\
\rowcolor{ourrow} PubMed-Summary    & Constrained summarization         & 2010 & 2025 & 160,000 \\
\rowcolor{ourrow} USPTO             & CPC patent classification             & 2006 & 2024 & 190,000 \\
WildTime-arXiv         & arXiv category classification        & 2007 & 2022 & 2,057,952 \\
\bottomrule
\end{tabular}
\end{table*}

\subsection{Recent Evaluations Better Predict Future Performance}
\label{subsec:frontier-evals}

Benchmarks collapse performance into a time-agnostic score, even when the data span many years, which obscures the temporal structure observed in \Cref{subsec:yearly-pretrain}. We thus evaluate GPT-5.6 Sol/Luna \citep{openai2026gpt56} on two popular benchmarks: (i) StreamingQA \citep{liska2022streamingqa}, a QA task based on news articles whose source date we take as the temporal axis, and (ii) SCIDOCA, a scientific citation prediction task, where publication date of the citing paper is taken as the timestamp.  In both cases, we group scores by year. 

\begin{figure}[t]
    \centering
    \includegraphics[width=\linewidth]{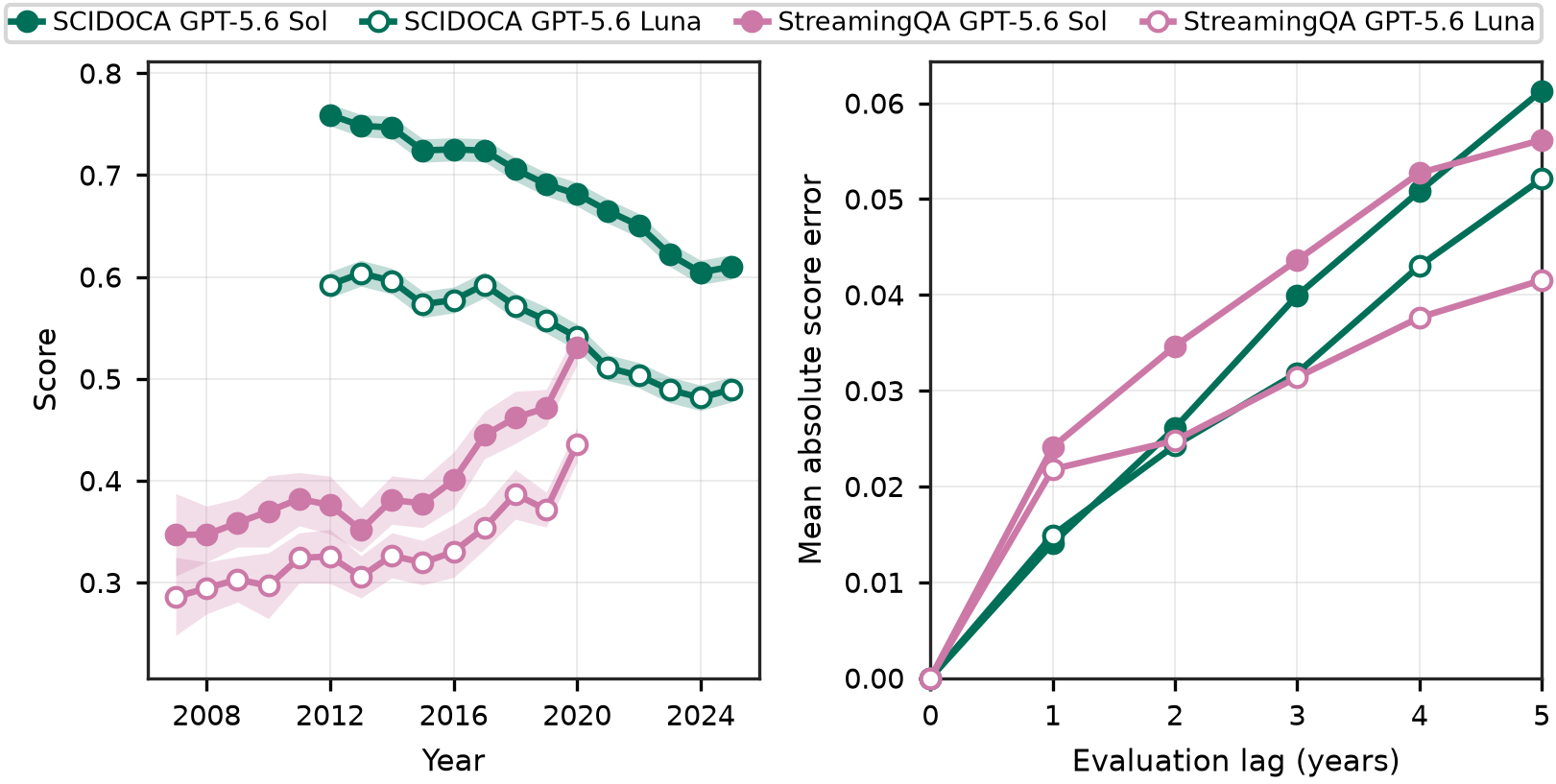}
    \caption{\textbf{Recent evaluations better predict future performance across different temporal performance trends.} We evaluate GPT-5.6 Sol/Luna on StreamingQA and SCIDOCA. \textit{(Left)} Depending on the task and model, performance changes over time in distinct ways. \textit{(Right)} Prediction error increases with evaluation age, so recent evaluations are the best predictors of future performance.}
    \label{fig:frontier-figure}
\end{figure}

We find that performance changes substantially over time on both datasets, but in opposite directions (\cref{fig:frontier-figure}, left). For both GPT-5.6 models, performance declines on SCIDOCA and improves on StreamingQA. Thus, even for a fixed model, a benchmark score depends meaningfully on the time period represented by the evaluation data.

Regardless of whether performance increases or decreases over time, recent performance better predicts future performance (\cref{fig:frontier-figure}, right). We frame this as a forecasting problem. For each model and dataset, we use the score from year $t-\Delta$ to estimate performance in year $t$, and call this the \textbf{$\Delta$-lag yearly predictor}. We see that the predictive error of the $\Delta$-lag predictor increases monotonically with evaluation age $\Delta$. Across all model-dataset pairs, the most recent evaluation is the best single-year predictor of next-year performance.

\begin{figure}[t!]
    \centering
    \includegraphics[width=\linewidth]{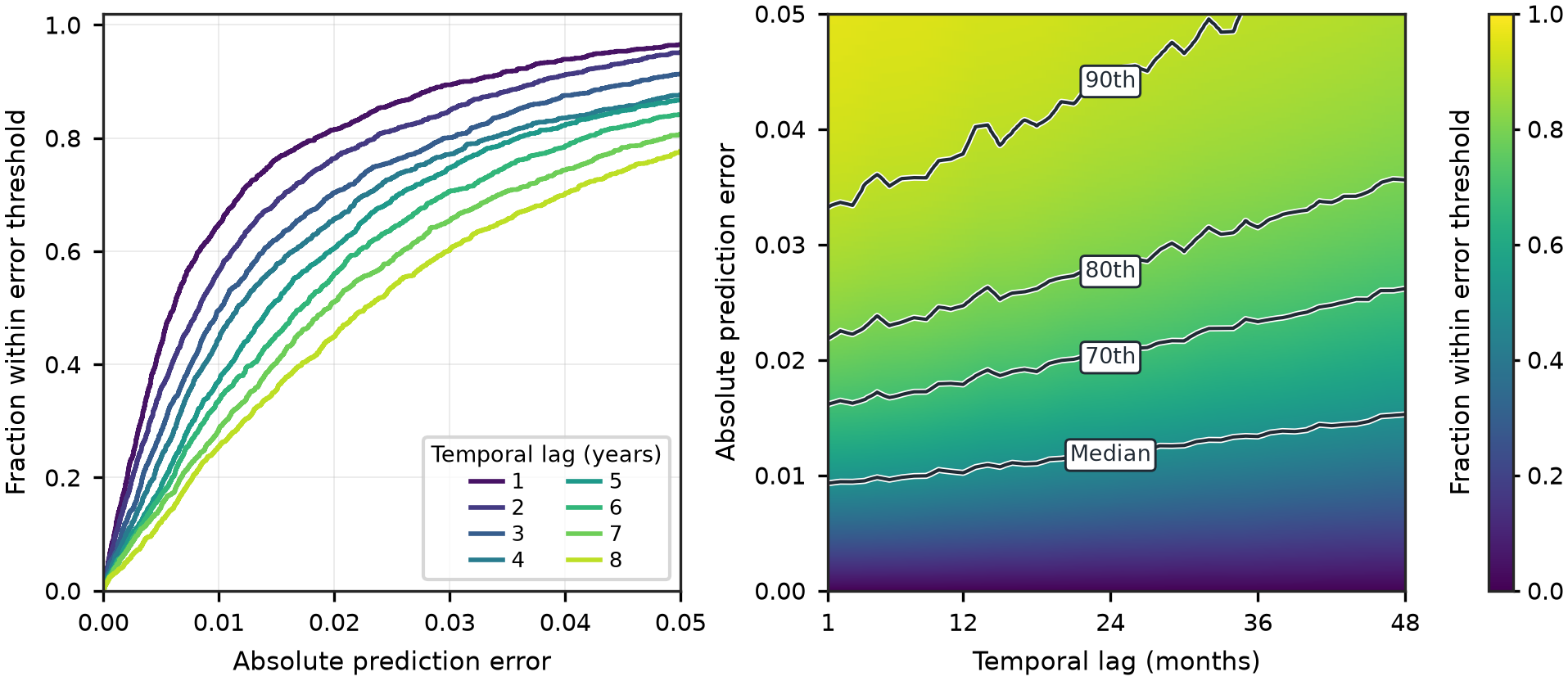}
    \caption{\textbf{Evaluation staleness smoothly degrades prediction quality at annual and monthly resolutions.} We evaluate 19 open-source models on 12 temporal datasets. \textit{(Left)} Annual prediction error CDFs shift downward as lag increases from 1 to 8 years. \textit{(Right)} Monthly prediction error rises as lag increases from 1 to 48 months. Error quantiles rise with lag, with the strongest degradation in the upper tail. Each comparison is restricted to periods where all plotted lags exist.}
    \label{fig:smooth-degrade}
\end{figure}

These observations show how stale evaluations grow less predictive for a few tasks and models, but do not establish whether this pattern persists more broadly across varying architectures, scales, and domains. We validate this pattern at scale using 19 open-source models from the Qwen3 \citep{yang2025qwen3}, OLMo 2 \citep{olmo20242}, Llama 3.1 \citep{grattafiori2024llama}, and Gemma 3 \citep{team2025gemma} families on 12 temporally structured datasets, evaluating yearly and monthly\footnote{We exclude FinQA and BioASQ from monthly analysis due to poor date recovery or very sparse coverage.} performance. 

At yearly resolution, increasing the evaluation staleness $\Delta$ monotonically shifts the error distribution of  $\Delta$-lag yearly predictors towards larger errors across models and datasets (\cref{fig:smooth-degrade}, left). This ordering holds across error thresholds, so the advantage of recent evaluations is robust across varying error tolerances. At monthly resolution, the predictive accuracy of \textbf{$\Delta$-lag monthly predictors} also degrades smoothly with age $\Delta$, rather than changing abruptly at some specific age (\cref{fig:smooth-degrade}, right). Additional analyses in \cref{appendix:wasserstein} support this conclusion. The quantile curves also show heightened degradation in the upper tail, so stale evaluations disproportionately increase large forecasting errors. 

This pattern extends to modern coding benchmarks such as LiveCodeBench \citep{jain2025livecodebench}, where next-month prediction error more than doubles as evaluation age increases from 1 to 12 months, and older monthly evaluations again increase the risk of large forecasting errors (\cref{appendix:lcb}). Such degradation persists even after controlling for model variants and unequal monthly sample sizes.

\subsection{Time-Aware Evaluation Better Predicts Future Performance}
\label{subsec:forecasting}

Because benchmark scores are used to inform expectations about future model performance, benchmark evaluation is already implicitly a forecasting problem. For each target year or month, forecasters receive only evaluations from earlier periods and predict performance in the next period. We represent time-agnostic pooling with a historical mean that ignores when evaluations were collected. \cref{subsec:frontier-evals} suggests that discarding such temporal structure can degrade predictive accuracy.

We compare the historical mean baseline with \textbf{persistence}, which predicts performance using only the most recent observation, equivalent to our 1-lag yearly or monthly predictor. We then test whether explicitly modeling the trajectory improves prediction\footnote{We restrict our analysis to target periods with a contiguous history of at least 5 years or 12 months.}, using the Chronos-2 time series foundation model \citep{ansari2025chronos} and four classical methods: linear extrapolation, ARIMA selected by AICc \citep{hurvich1989regression}, Simple Exponential Smoothing (SES) \citep{holt1957forecasting}, and damped Holt \citep{gardner1985forecasting}. In the left panel of \cref{fig:main-figure}, we plot error CDFs of selected annual forecasters; all annual and monthly forecasters and their MAEs are in \cref{appendix:all-forecasters}.

At annual and monthly resolutions, persistence reduces MAE relative to the historical mean by roughly \emph{one-third and one-fifth}, respectively, representing a drastic improvement over standard practice. At monthly resolution, \emph{every explicit time series forecaster improves further upon persistence}. 

We observe even greater recency benefits on LiveCodeBench, where persistence reduces MAE by 45.1\% relative to the historical mean. This advantage is robust to alternative historical weighting, model variants, evaluation months, unequal monthly sample sizes, and training cutoffs (\cref{appendix:forecast-lcb}).

\begin{findingbox}
\textbf{Observation 3: %
Time-aware forecasters substantially outperform the standard practice of computing a uniform averaging over all historical data.} %
\end{findingbox}

\subsection{Time-Aware Forecasters Improve Model Selection}
\label{subsec:model-ranking}
 
Above, we found that time series forecasters can substantially improve predictions of future model performance. These forecasts support various deployment choices, such as determining if a model is likely to clear a task-specific performance bar, or choosing among candidate models. 
\cref{appendix:inversions} shows \emph{ranking inversions}, i.e. cases where one model outperforms another in one evaluation period but underperforms it in another, between thinking and non-thinking variants, different model sizes in a family, similarly capable models across families, and frontier versus open-weight models.

We now ask whether time series forecasters improve modeling and deployment choices. We focus on model selection: predicting which model from a candidate set will perform best on new data. We give a forecaster $K$ models and select the model it predicts will achieve the highest performance on new data. We measure how often this procedure selects the model that actually performs best in the target period, averaged over all subsets of size $K$ drawn from 19 total models and pooled across benchmarks and target periods. We call the resulting measurement the \textbf{optimal selection rate}.

The optimal annual selection rate for any method naturally degrades as the shortlist grows, but time series forecasters perform significantly better than historical pooling (\cref{fig:main-figure}, right).\footnote{We provide a concrete failure case of historical pooling on WildTime-arXiv \citep{yao2022wild} in \cref{appendix:inversions}.} At monthly resolution, Chronos-2 also improves model selection over historical pooling (\cref{fig:full-selection}, left).\footnote{Persistence does not improve monthly model selection, despite reducing monthly forecasting MAE. Much of the forecasting gain comes from errors shared across models, which do not affect their rankings (\cref{appendix:forecast-vs-select}).}

Optimal selection rate treats all mistakes equally, even though selecting a near-tied model is less consequential than selecting one that performs far worse. This motivates us to consider selection \textbf{regret}, the accuracy gap between the selected and optimal models. We define high-regret choices as those exceeding two accuracy points. Annual time-aware methods roughly halve the rate of high-regret choices relative to the historical mean baseline (\cref{fig:full-selection}, right). Thus, time-aware evaluation can reduce not only forecast error but also the frequency of high-regret model selections. 

\begin{findingbox}
\textbf{Observation 4: Time-aware evaluation improves model selection.}
Model selection should account for when evaluation data were collected, whether by weighting recent samples higher or by applying a time series forecaster to extrapolate historical evaluation trends.
\end{findingbox}

\section{Temporal Recency Impacts Training Data Selection}

We now ask whether temporal shifts should inform how practitioners select training data. Pretraining pipelines already optimize data via deduplication, quality filtering, source and domain balancing, and stage-specific mixtures that trade off diversity and quality \citep{olmo2025olmo,blakeman2025nvidia}. We argue that \emph{recency} should also be considered in data filtering and sampling rates. 

\textbf{We first ask whether restricting pretraining to recent data improves near-future performance.} Extending the experiments in \cref{subsec:yearly-pretrain}, at each cutoff we pretrain on all FineWeb data up to that cutoff (cumulative) or just the last five years (recent-window), comparing their next-year perplexities on FineWeb, PubMed, and arXiv across five training recipes\footnote{We use the five learning rate and weight decay configurations from 1B-parameter tuning in DCLM.}. On FineWeb and PubMed, recent-window usually has lower next-year perplexity than cumulative, whereas arXiv shows no clear advantage from pretraining on recent data (\cref{appendix:training-data}). Prior work has similarly shown that replaying older data can improve retention, and the optimal balance between new and old data is domain-dependent \citep{li2025tic}. Thus, we do not suggest discarding old data. Instead, practitioners could decrease sampling rates with age while retaining historical coverage.

While the appropriate level of recency depends on the target domain, the most recent yearly evaluation selects the better model between recent-window and cumulative in 97\% of cases across all datasets. Furthermore, sensitivity to evaluation age is domain-dependent: on FineWeb, selection accuracy falls to 51\% with five-year-old evaluations, while PubMed and arXiv selection accuracy remain stable across lags (\cref{appendix:training-data}). For data mixture search methods such as DCLM and NVIDIA CLIMB \citep{li2024datacomp,diao2026nemotron}, these two observations suggest adding recency to the search space and using recent validation on the target domain to select among candidate mixtures.

\textbf{We next ask whether the same principle extends to late-stage training.} Here, practitioners deliberately reshape data mixtures: OLMo 3 curates high-quality math, science, code, and instruction-following data \citep{olmo2025olmo}, Qwen3 increases STEM and reasoning data \citep{yang2025qwen3}, and continued pretraining (CPT) recipes optimize the composition of adaptation data \citep{parmar2024reuse}. We take each pretrained checkpoint\footnote{We use the highest learning rate recipe for each cumulative and recent-window model. The recipe rankings are almost constant across evaluation years, suggesting that temporal sensitivity is not due to poor optimization.} and perform CPT on 1.6M tokens of PubMed or arXiv data from each year of the pretraining window. In both domains, five-year-old CPT data perform worse than current-year CPT data at every cutoff, making the most recent CPT data a strong baseline.

Recent adaptation data typically improve next-year performance, but the benefit varies by domain: the most recent year is always best for PubMed, but only in 12 of 18 arXiv windows. On arXiv, selecting the candidate with the best current-year performance reduces mean next-year selection regret by 87\%. Thus, practitioners should preserve timestamps and treat adaptation year as a data selection variable, using the most recent data when performance is ordered by adaptation age, and recent target-domain evaluation otherwise.

\textbf{Finally, we ask whether recency matters when selecting expert-labeled data for SFT.} We simulate a fixed budget for cybersecurity expert annotations on NVD-RCM, finetuning Llama-3.1-8B-Instruct on examples sampled either from all preceding years or only the most recent year. Sampling recent examples improves mean next-year accuracy in each tested year and budget, by 2.63 percentage points on average (\cref{appendix:active-learning}). This setting is closely related to active learning, which selects unlabeled examples for annotation under a constrained budget \citep{settles.tr09, bhatt2024experimental}. Our results suggest that even before applying more sophisticated data selection methods, practitioners should use recency to help prioritize which examples experts annotate.

\begin{findingbox}
\textbf{Observation 5: %
Recent data are often preferable for pretraining and continued pretraining, and for SFT under an annotation budget.} When the best temporal policy is unclear, recent target-domain evaluation can provide strong guidance.
\end{findingbox}

\section{Discussion}
\label{sec:discussion}

Our results suggest that LLM evaluation should be treated as a forecasting problem. Recent evaluations predict future performance better than standard pooled benchmarks, and time-aware forecasters further improve prediction and model selection. In deployment, accounting for temporal structure reduces prediction errors and high-regret model choices. This perspective also informs training data selection, where recency can improve near-future performance but is not universally optimal. 

More broadly, we recommend that practitioners preserve temporal information, so that evaluation and data selection reflect the evolution of data distributions and model scores over time. In current practice, benchmark scores are often pooled into a single time-agnostic metric, while training data mixtures are optimized for factors such as domain and quality without directly accounting for date age. Our results suggest that practitioners should treat time as an explicit axis of both evaluation and data selection, prioritizing recent data or employing target-domain validation when appropriate. 

Future work could use temporal structure to decide when to reevaluate models and refresh training data. Rather than updating benchmarks at fixed intervals, evaluation systems could determine when existing scores grow stale, and then gather data that are more informative about future performance or model selection. Similarly, training data selection could balance the benefits of recent data against its collection and labeling costs, especially when annotation requires domain expertise.

\newpage

\ifdefined\ARXIVVERSION
\else

\section*{AI use statement}

In this work, we used generative AI tools to polish writing and figures, search for related work, and search for source corpora with sufficient yearly coverage for our temporal analysis. We have not used generative AI tools to perform any research ideation or execution, to draft sections of the paper, or to generate synthetic datasets; proving mathematical claims is not applicable to this work. We have reviewed all AI-assisted work. For example, we checked all LLM-generated corpora suggestions through a manual literature survey, and we carefully verified relevant data processing techniques from prior work and cited any such works in \cref{appendix:datasets}. We take responsibility for the final content of this work, including text, claims or artifacts produced with the aid of generative AI.

\section*{Reproducibility statement}

We include full details to reproduce our dataset constructions from source corpora in \cref{appendix:datasets}, as well as full details on dataset scoring and model inference in \cref{appendix:model-generation}. We further include full details for our yearly and multi-year pretraining experiments in \cref{appendix:training-details}.

\fi

\bibliography{iclr2027_conference}
\bibliographystyle{iclr2027_conference}

\newpage

\appendix

\crefname{section}{Appendix}{Appendices}

\section{Related Work}
\label{appendix:related-work}

\paragraph{Temporal distribution shift and adaptation.} Prior work has shown that text distributions and model performance shift over time. \citet{lazaridou2021mind} show that language models degrade on text beyond their cutoff, while \citet{yao2022wild} benchmark natural temporal distribution shifts across many domains. \citet{cheng2024dated} further show that a model's effective knowledge cutoff wildly varies across data sources. Other works explicitly adapt models over time: TemporalWiki uses continual updates on evolving knowledge \citep{jang2022temporalwiki}, TiC-LM uses continual pretraining on 114 Common Crawl dumps \citep{li2025tic}, \citet{nylund2024time} derive temporal model updates that generalize to the future, and \citet{pilchen2026understanding} find that chronological ordering improves factual freshness in pretraining. Our work instead provides a unified framework connecting temporal pretraining data shifts, evaluation staleness, forecasting, model selection, and training data selection. 

\paragraph{Time-aware evaluation and model selection.} Some benchmarks incorporate time through timestamped or continually refreshed evaluation data \citep{liska2022streamingqa,vu2024freshllms,white2025livebench,jain2025livecodebench}. Temporal information has also been used as a signal of benchmark contamination. \citet{roberts2023data} compare model performance on problems before and after training cutoffs, while \citet{zhang2026temporal} show that post-cutoff performance decay depends on benchmark construction. We use temporal information to study a different question: how does the age of an evaluation impact its ability to predict future performance? Most closely related to our work, \citet{han2024model} studies model assessment and selection under temporal distribution shift, using adaptive rolling windows to estimate \emph{current} generalization error from current and historical data. We instead treat benchmark evaluation as a forecasting problem, using temporally ordered evaluations to predict performance in the \emph{future}, and studying how model selection changes as evaluations age.

\paragraph{Training data selection and continued pretraining.} Prior work has studied how training data composition impacts model performance. Domain-adaptive pretraining improves performance on a target domain \citep{gururangan2020don}, DSIR selects pretraining data to match a target distribution \citep{xie2023data}. Later work on late-stage training studies domain upsampling \citep{blakeney2024does} and data selection for continued pretraining \citep{parmar2024reuse}. Time-continual approaches further study how models should incorporate new data while retaining older knowledge \citep{li2025tic}. We complement these approaches by treating \emph{data recency} as an explicit design variable and using recent, time-forward validation to select  training data policies for near-future performance.

\newpage

\section{Additional Temporal Shifts in Pretraining Corpora}
\label{appendix:additional-metrics}

\begin{figure}[h]
    \centering
    \includegraphics[width=\linewidth]{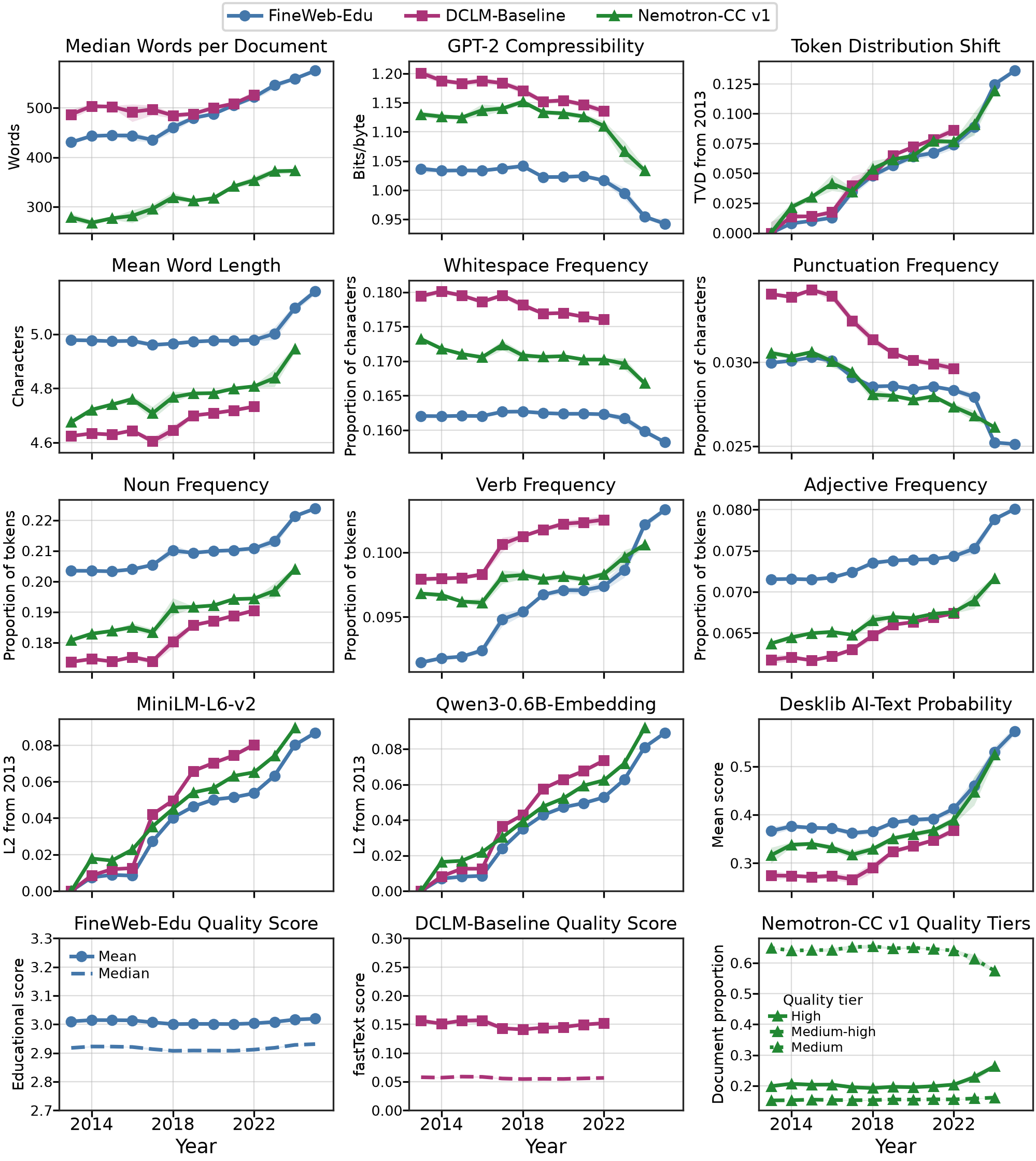}
    \caption{\textbf{Additional corpus metrics show systematic temporal distribution shift across pretraining corpora.} FineWeb-Edu, DCLM-Baseline, and Nemotron-CC v1 exhibit changes in document statistics and token and part-of-speech composition \textit{(first three rows)}. Annual document-embedding centroids shift over time, while estimated AI-text probability rises sharply in recent years \textit{(fourth row)}. Dataset-specific quality signals remain relatively stable \textit{(fifth row)}.}
    \label{fig:pretraining-trends-more}
\end{figure}

\Cref{fig:pretraining-trends-more} shows additional diagnostics supporting the temporal distribution shifts in pretraining corpora. As embedding-centroid distances are difficult to interpret in absolute terms, we compare them with distances between corpora. Temporal drift within a single corpus reaches roughly the full 2013 DCLM-Nemotron distance and half the 2013 FineWeb-DCLM distance, showing that within-corpus temporal variation can rival the differences between distinct corpora. This drift largely reflects a shift in embedding location rather than local density: for both Qwen encoders, changes in mean 10th-nearest-neighbor distance over time are less than 10\% of the corresponding centroid shift. The figure also reports additional document, token, AI-text, and dataset-specific quality metrics.

\newpage

\section{Training Details}
\label{appendix:training-details}

We follow the standard DCLM architecture \citep{li2024datacomp}. Each model has 411,665,408 trainable parameters with 24 pre-norm transformer layers. Each layer has width 1024, contains 8 attention heads of dimension 128, and contains a 2,816-dimensional bias-free MLP with SwiGLU. As in DCLM, we use RoPE, QK-normalization, LayerNorm, untied input and output embeddings, and no dropout. The tokenizer is GPT-NeoX. Each model is trained on exactly 4,019,200 packed sequences of length 2,048 (a total of 8,231,321,600 tokens), using 7,850 optimizer steps with a global batch size of 512 sequences. Documents are separated by EOS and packed without padding, following the standard DCLM convention. Evaluation is carried out on 32,768 fixed-length sequences, for a total of 67,108,864 tokens. In training, we use AdamW with $\beta_1=0.9,\beta_2=0.95,\epsilon=10^{-8}$. We use a linear warmup for the first 2000 steps to the peak learning rate, then cosine decay to $3\times 10^{-5}$. The gradient norm is clipped to $1$, and we include a $10^{-4}$ $z$-loss term. We conduct training on NVIDIA B200s, and we use BF16 autocast and seed 124. 

Continued pretraining uses the exact same settings, with the warmup step count proportionally scaled to the token count. For the NVD-RCM SFT experiment on Llama-3.1-8B-Instruct, we use LoRA (\(r=8,\alpha=16\)) for one epoch with learning rate \(3\times10^{-4}\), global batch size 16, and otherwise hold the SFT recipe fixed across conditions.

\newpage

\section{All Excess Loss Heatmaps}
\label{appendix:yearly-heatmaps}

\begin{figure}[h]
    \centering

    \includegraphics[width=0.57\textwidth]{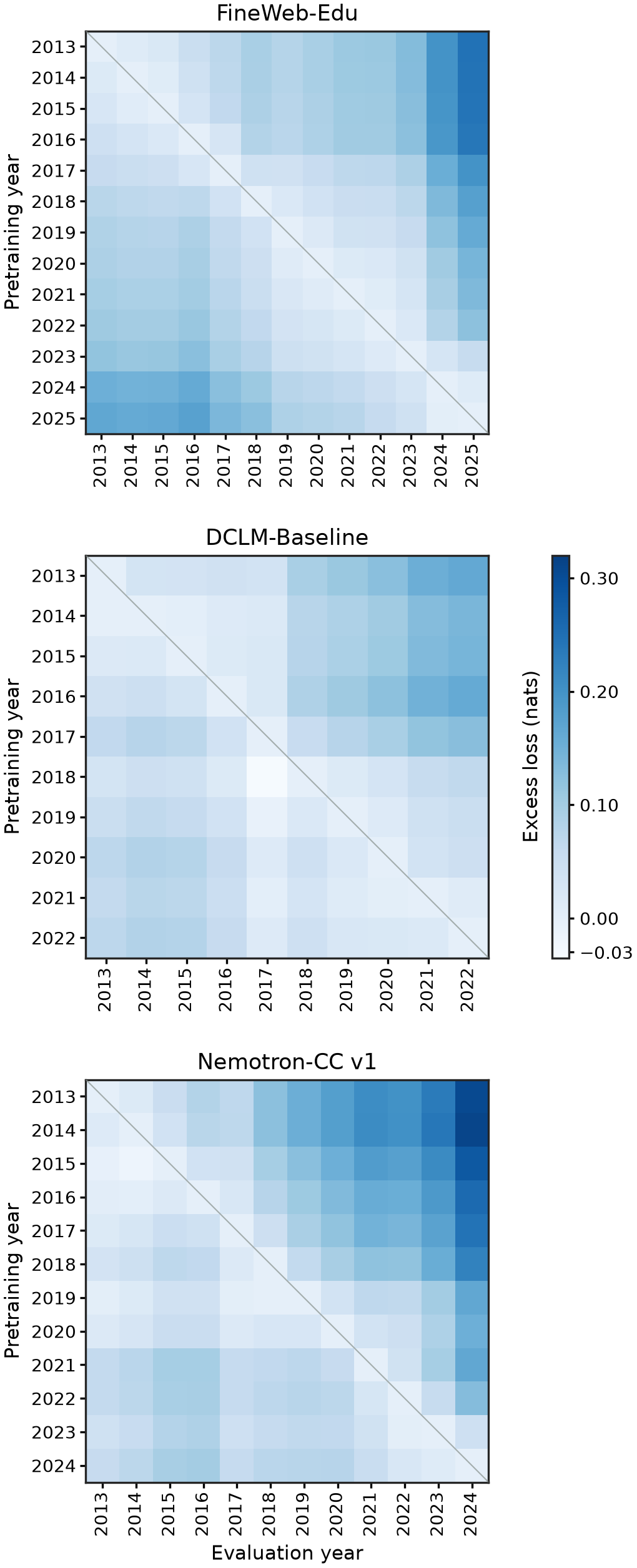}

    \caption{\textbf{Models generalize best to nearby years across all three pretraining corpora.} On FineWeb-Edu, DCLM-Baseline, and Nemotron-CC v1, excess loss $\Delta L(s,t)=L(s,t)-L(t,t)$ is smallest near the diagonal and increases as the pretraining year $s$ and evaluation year $t$ separate.}
    \label{fig:three-heatmaps}
\end{figure}

\newpage

\section{Datasets}
\label{appendix:datasets}

\subsection{BioASQ Task B}

We construct a temporal biomedical QA benchmark from BioASQ Task B \citep{krithara2023bioasq}, using the official Golden Enriched test sets from BioASQ 2B through 13B (2014--2025). Phase B provides gold supporting snippets, so our task is context-grounded QA rather than closed-book QA. We process each annual set independently using the SciRIFF BioASQ preprocessing and instruction templates \citep{wadden2025sciriff}, preserving the original annual splits instead of using SciRIFF's repartitioning. We use template 0 with its 4,096-token configuration and drop overlength examples. SciRIFF concatenates gold snippets into the context, maps factoid synonym lists to the longest answer string, and keeps the first synonym for each entity in list questions. We exclude summary questions because they require judge-based evaluation. Of the 5,486 raw questions, we retain 4,067 yes/no, factoid, and list questions with gold answers.

Models receive the question with concatenated gold snippets and must generate an answer in the format required by the question type. Yes/no responses are scored by exact accuracy after stripping whitespace and lowercasing. Factoid responses use exact match against the single SciRIFF-converted reference after lowercasing, removing punctuation and articles, and collapsing whitespace. List responses are parsed with SciRIFF-style JSON extraction, normalized with SciRIFF's list normalization and compared as sets using exact-entry F1 scores, with unparseable responses receiving a score of zero. The annual score is the mean per-question score across all examples.

The temporal axis is the challenge year. We do not perform monthly analysis because each year's test questions are concentrated in a short challenge window, making monthly evaluation highly sparse.

\subsection{StreamingQA}

We construct a temporal evolving-knowledge QA benchmark from StreamingQA \citep{liska2022streamingqa}, which is built from English WMT News Crawl articles. We start from the official 36,378-example evaluation file and retain only examples with \texttt{recent\_or\_past = past}, as these span the historical corpus rather than clustering at 2020. This gives 16,307 examples from August 2007 to December 2020. We preserve the original question, its three official reference answers, the question timestamp, and the source-article timestamp.

Models are evaluated with closed-book QA. Following StreamingQA, each question is prefixed with "Today is \{date\}" using the question timestamp. We append a fixed instruction requesting only the short answer. Predictions are scored with SQuAD-style token F1 \citep{rajpurkar2016squad}: prediction and reference strings are lowercased, ASCII punctuation and articles are removed, and whitespace is collapsed. We compute token F1 against each of the three official references and keep the maximum. The cohort score is the mean per-question F1. 

The temporal axis is the source-article timestamp, not the question timestamp. Monthly analysis uses the source-article calendar month directly, requiring no additional metadata or matching. The full span contains 161 months, of which 151 have data and 10 are represented as missing observations.

\subsection{FinQA}

We construct a temporal financial numerical-reasoning benchmark from FinQA \citep{chen2021finqa}. We combine train, development, and test splits to get 8,281 examples. FinQA uses S\&P 500 earnings reports from 1999--2019. We keep all examples and assign report years from the canonical ID. 

For each example, the model receives the complete \texttt{pre\_text}, the full table rendered in Markdown, the complete \texttt{post\_text}, and the question, all in source order. The Markdown conversion preserves table structure while also handling irregular row lengths and cell formatting. Gold supporting facts, programs, execution results, reasoning steps, and answers are not included in the input. 

FinQA scoring is more involved because of heterogeneous answer formatting. For parseability, we request a final answer in the form "Final answer: ...". For scoring, \texttt{qa.answer} is used if it is yes/no or contains exactly one percentage quantity; otherwise it must contain exactly one number. If unusable, the same rule is applied to \texttt{qa.steps[-1].res}. Predictions are extracted from the final answer marker, then conclusion phrases, and then last-line fallbacks. Parsing supports signed and comma-separated numbers, currency, percentages, leading decimals, accounting parentheses, and ratios and fractions, but never computes unfinished arithmetic. Numeric predictions are compared after rounding both prediction and reference to the number of decimal places shown in the reference, and percentage and non-percentage values must also match in type. Ambiguous outputs score zero.

We do not perform monthly analysis on FinQA because the processed benchmark only gives report years and it is difficult to externally reconstruct all filing dates.

\subsection{WildTime-arXiv}

We use the arXiv dataset from Wild-Time \citep{yao2022wild}, a 172-way classification task that predicts an arXiv paper's primary subject category from its title. It includes 2,057,952 examples from 2007--2022, with the temporal axis being the arXiv metadata field \texttt{update\_date} rather than initial submission year. Models receive the paper title and the fixed list of 172 valid category codes, and must return exactly one code. We score 172-way classification accuracy: a response is correct if it contains exactly one valid category from the Wild-Time label set and matches the gold label.

We download the official \texttt{arxiv.pkl} from the Wild-Time repository, which retains lowercased titles and integer class IDs but not category strings. We recover the 172 category strings by matching titles against Cornell's \texttt{arxiv-metadata-oai-snapshot.json} \citep{cornell_arxiv_dataset}. We verify that the recovered mapping reproduces Wild-Time’s class ID ordering. Because the arXiv taxonomy changes over time, we preserve Wild-Time's original label space instead of canonicalizing labels. We restore three historical labels from the original Wild-Time label space, replacing current labels \texttt{math.NA}, \texttt{eess.SY}, and \texttt{econ.GN} with \texttt{cs.NA}, \texttt{cs.SY}, and \texttt{q-fin.EC}.

For monthly analysis, we recover month-level dates from the same metadata. Each Wild-Time example is matched by its title to a current arXiv record, and we accept its \texttt{update\_date} only if its year agrees with the official Wild-Time year. We discard examples without a unique valid match. Yearly analysis retains all examples, while monthly data retain 94.3\% of original examples.

\subsection{PubMed}

We construct PubMed-Summary, PubMed-MeSH, and PubMed-Rhetorical from the 2026 PubMed Baseline (39,994,988 records) released by NLM \citep{nlm2026pubmed}. We assign each record to the earliest year available in \texttt{PubDate} or \texttt{ArticleDate} and retain 2010--2025. Summary and MeSH share the same 160,000 English MEDLINE records with nonempty abstracts and at least one MeSH descriptor, sampled at 10,000 random records per year. Rhetorical independently samples 10,000 English MEDLINE randomized controlled trials per year, requiring 3--9 structured-abstract sections whose publisher labels map to the five PubMed-200k-RCT roles \citep{dernoncourt2017pubmed, ripple2015structured}.

Summary provides the title and abstract and requests exactly three bullet lines totaling at most 50 words. Each example receives a binary instruction-following score for satisfying both constraints. MeSH provides the title and abstract and requests a JSON string array of MeSH descriptors. Predictions are mapped to descriptor IDs using the 2026 MeSH vocabulary, and each article is scored with set F1. Unresolved strings are false positives and invalid arrays are empty predictions. Rhetorical sentence-segments each structured abstract with Stanford CoreNLP \citep{manning2014stanford}, assigns each sentence its section role, and removes section headings that leak the label, giving 1,978,916 sentences. Models get numbered sentences and predict one of five rhetorical roles for each, and performance is mean per-abstract sentence accuracy.

For monthly analysis, we rejoin sampled PMIDs to the same baseline XML. We assign a month only when the publication month is unambiguous and consistent with the annual cohort. This retains 142,553 of 160,000 Summary/MeSH and 146,950 of 160,000 Rhetorical examples in 192 months.

\subsection{USPTO}

We construct a temporal patent-classification benchmark similar to USPTO-70k \citep{pujari2021multi}. The task is to predict four-character Cooperative Patent Classification (CPC) subclasses from U.S. patent titles and abstracts. We use the final 2024 PatentsView metadata release \citep{uspto_patentsview_2025} and follow the construction from \citet{pujari2024neural}, aggregating current CPC assignments by patent and keeping all CPC subclasses. For each grant year from 2006--2024, we keep utility patents with non-empty titles and abstracts and at least one CPC subclass. This yields 5,152,802 eligible patents, and we sample 10,000 patents per year to obtain 190,000 total examples.

Models receive the patent title and abstract and output up to seven four-character CPC subclasses in JSON format, like in \citet{emer2026large}. We normalize predicted labels to uppercase four-character codes, deduplicate them, and treat unparseable outputs as empty. Each example is scored with set F1 against the full set of gold CPC subclasses, and the cohort score is the mean across examples.

The temporal axis is the patent grant date from PatentsView. Monthly analysis uses the grant month.

\subsection{NVD}

We construct temporal versions of CTIBench's Root-Cause Mapping (RCM) and Vulnerability-Severity Prediction (VSP) tasks \citep{alam2024ctibench}. Because NVD yearly feeds are partitioned by CVE-ID prefix and not publication date, we pool all feeds before forming temporal cohorts. RCM predicts a CWE identifier from a CVE description, while VSP predicts its CVSS v3.1 base vector. We use AthenaBench's NVD filtering and prompt templates \citep{alam2025athenabench}.  We use the official NVD JSON 2.0 feeds for 2002--2025 \citep{booth2013national} and apply AthenaBench-style filtering. For RCM, we keep English descriptions with exactly one valid CWE label and at least 25 words. For VSP, we require a CVSS v3.1 record, prefer NVD-authored metrics over CNA metrics when available, and require at least 30 words. We globally deduplicate descriptions, retaining 132,103 RCM examples from 2017--2025 and 116,711 VSP examples from 2019--2025.

We use the official AthenaBench prompt templates. RCM requests a short justification followed by a CWE identifier and is scored with exact CWE accuracy. VSP requests the eight CVSS base metrics and a final CVSS vector. An example is correct if all eight predicted components match the gold vector, with incomplete predictions scored as zero. This exact-vector score differs from AthenaBench, which evaluates VSP using CVSS-score deviation.

The temporal axis is the NVD \texttt{published} timestamp, not the CVE-ID year or \texttt{lastModified}. Monthly analysis uses the publication month directly, requiring no external metadata.

\subsection{SCIDOCA}

We construct SCIDOCA from SCIDOCA 2025 Task 2 \citep{dao2025overview}, using the frozen Semantic Scholar 2026-09-01 S2ORC and metadata release \citep{lo2020s2orc,kinney2023semantic}. Source papers must contain body text and citation annotations that can be linked to cited papers. We assign each citation to the paragraph that contains it, and keep only paragraphs where all citations are resolved and at least one non-cited candidate paper is available as a distractor. Citations are replaced by \texttt{[CITATION\_i]} masks. Following SCIDOCA, the candidate set for each example includes all distinct papers cited anywhere in the source paper.

We join candidate titles and abstracts from Semantic Scholar metadata. Candidates lacking either are removed, as are citation masks whose gold papers lack complete metadata. Examples must retain at least one gold citation and one distractor. Within each year, we rank eligible source papers, consider up to the first 15,000 before metadata filtering, and retain the first 5,000 surviving examples. This produces 70,000 examples from distinct source papers spanning 2012--2025.

Models receive the masked paragraph and each candidate's CorpusId, title, and abstract, and return a JSON \texttt{predicted\_citations} mapping. Predicted CorpusIds are compared to the gold mapping after simple string normalization. Per-example accuracy is the fraction of citation masks predicted correctly, and cohort scores are simple averages of these accuracies.

The temporal axis is the citing paper's publication date. For monthly analysis, we recover publication dates from Semantic Scholar metadata. Dates are available for 96.9\% of examples and span all 168 months from 2012-01 through 2025-12.

\subsection{CLERC}

We construct a temporal legal precedent-identification benchmark from CLERC using its CAP-derived corpus \citep{hou2025clerc}, with decision dates and citation metadata recovered from the Harvard Caselaw Access Project (CAP) \citep{cap}. Each example pairs an indirect single-removed (ISR) query with a candidate case. Following CLERC, queries use 150 words on each side of a central citation. The central citation and salient sentence are masked and direct-quotation contexts are excluded. We recover CAP dates and citations by scanning U.S.-jurisdiction volumes and exact-matching CAP IDs to CLERC, recovering 1,838,630 of 1,842,422 documents. We normalize reporter citations using CLERC's \texttt{clean\_cstr} procedure and keep only aliases that uniquely identify a CLERC document. Citation extraction follows CLERC's eyecite-based code \citep{cushman2021eyecite}.

From 2006--2020, we retain at most one query per cited target in each source case, and exclude targets decided after the source case, yielding 2,691,044 eligible citation events. We drop 2020 as it is incomplete. For each year from 2006--2019, we randomly sample 5,000 eligible citation events. Each event yields one positive example containing the cited case and one hard negative, defined as the highest-ranked BM25-retrieved candidate that is neither the gold nor source case and was decided before the source case. We retrieve over the full corpus using Pyserini 0.24.0/Anserini with BM25 ($k_1=0.9,b=0.4$) \citep{lin2021pyserini,yang2017anserini}. The final benchmark contains 140,000 examples, with exactly 5,000 YES and 5,000 NO examples per year.

The model gets the masked passage and candidate text and must return YES/NO. We score accuracy after normalizing predictions to YES or NO, giving invalid outputs a score of zero. The temporal axis is the source case's CAP decision date. Monthly analysis uses the \texttt{source\_date} field.

\newpage

\section{Model Generation Settings}
\label{appendix:model-generation}

For closed-source models, we evaluate GPT-5.6 Sol and Luna \citep{openai2026gpt56} under deterministic, tool-free inference. Each benchmark example is supplied with its native prompt and a shared system instruction "You are a helpful assistant." Models receive no web access, retrieval, tools, or information beyond the benchmark. On all datasets, we use temperature 0, a maximum output of 16,384 tokens, no tools, and no response storage. We additionally set reasoning effort to none.

For open-source models, we use 13 instruction-tuned open-weight models spanning four families: Qwen3 \citep{yang2025qwen3} at 0.6B, 1.7B, 4B, 8B, 14B, and 32B; Gemma-3-IT \citep{team2025gemma} at 4B, 12B, and 27B; OLMo-2-1124-Instruct \citep{olmo20242} at 7B and 13B;
and Llama-3.1-Instruct \citep{grattafiori2024llama} at 8B and 70B. We use the same system instruction as GPT-5.6, with no tools, retrieval, or web access. We decode with temperature 0. For Qwen3, the only family in our panel whose chat template has a thinking toggle, we additionally evaluate each model with thinking on under the recommended sampling settings (temperature 0.6, top-$p$ 0.95, top-$k$ 20). With these 6 thinking variants, we obtain a total of 19 models.

Given the varying sizes and difficulties of our 12 tasks from \cref{appendix:datasets}, we set different maximum token settings. In general, we generate up to 32,768 total tokens for non-OLMo models and 4,096 total tokens for OLMo models (their native context length). For some models and tasks, we additionally impose limits on the number of maximum new tokens generated. For thinking variants, we cap generations to 4,096 new tokens on all datasets except BioASQ, StreamingQA, and FinQA, as longer reasoning trajectories are almost always repetitive or unparseable. For non-thinking variants, we set maximum new tokens based on task complexity for larger datasets: 4 for CLERC (a simple yes/no task), 256 for SCIDOCA (citation prediction), and 1,024 for WildTime-arXiv and both NVD tasks (arXiv categories or short cybersecurity answers). These thresholds were determined to be adequate by observing that the fraction of truncated responses remained negligible. The Fakeddit experiments in \cref{appendix:vlm-graphs} likewise used a maximum of 4 new tokens, due to short yes/no responses.

\newpage

\section{Temporal Locality in Modern Coding Benchmarks}
\label{appendix:lcb}

We use the public code-generation leaderboard from LiveCodeBench\footnote{ \url{https://github.com/LiveCodeBench/livecodebench.github.io}, accessed 10/03/26.} \citep{jain2025livecodebench}, a continuously updated benchmark that contains evaluations for 28 models on 1,055 problems from May~2023 to April~2025. We aggregate official per-problem \texttt{pass@1} scores per release month.

\begin{figure*}[h]
    \centering
    \includegraphics[width=\textwidth]{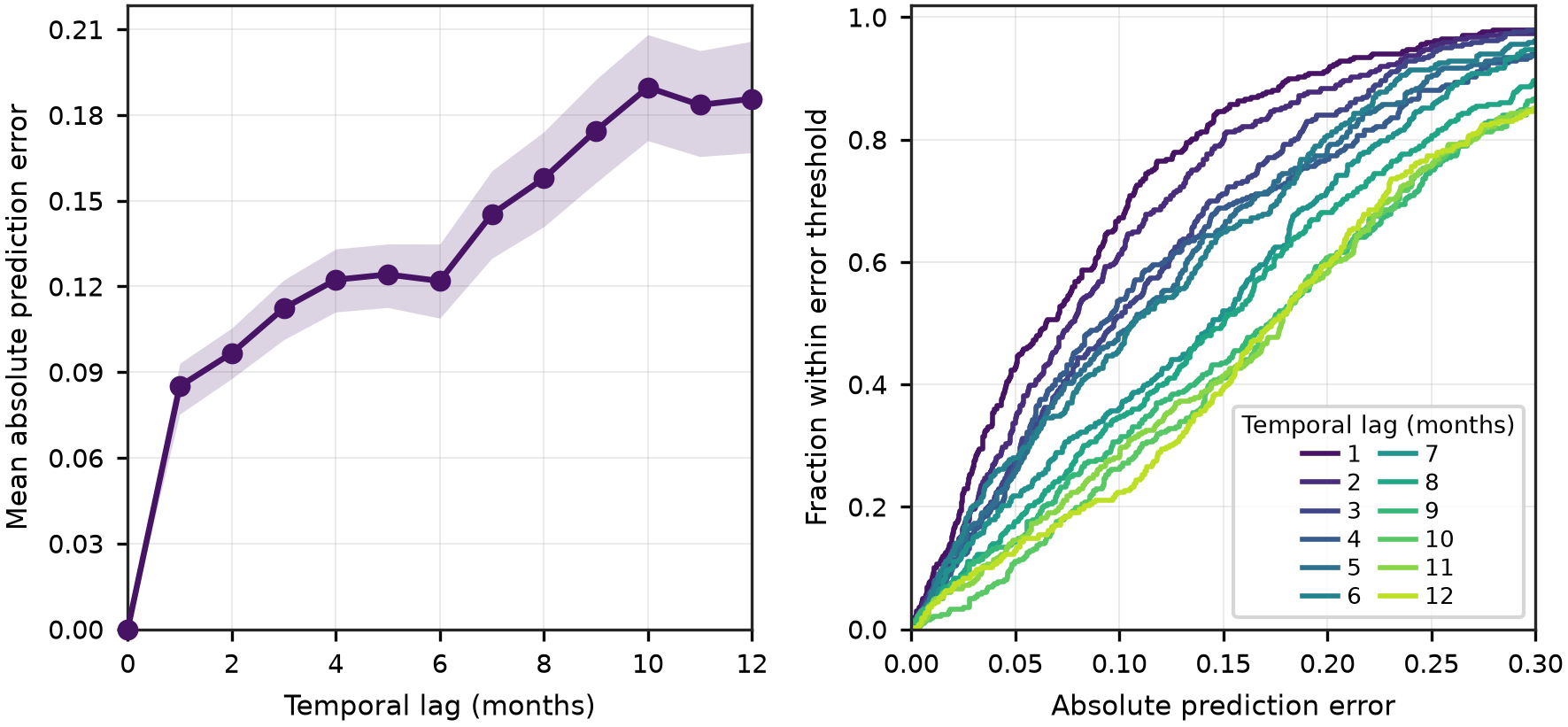}
    \caption{\textbf{LiveCodeBench evaluations become less predictive as they grow stale.} \textit{(Left)} Prediction error increases with monthly lag, so recent evaluations are the best predictors of future performance. Shaded bands are 95\% confidence intervals from resampling within model families. \textit{(Right)} Monthly prediction error CDFs shift downward as lag increases from 1 to 12 months. Each comparison is restricted to periods where all plotted monthly lags exist, giving 336 data points at every lag.}
    \label{fig:lcb_staleness}
\end{figure*}

We repeat our monthly lag analysis in \cref{fig:lcb_staleness}. Once again, the most recent monthly evaluation is substantially more predictive of next-month performance than older evaluations. This effect is not driven by over-representation of related model variants, as collapsing the 28 models into 20 base families gives a similar increase in MAE from 0.082 at monthly lag 1 to 0.187 at monthly lag 12. We also verify that this pattern is robust to equalizing the number of problems per month (\cref{appendix:forecast-lcb}).

\newpage

\section{Wasserstein Monthly Lag Analysis}
\label{appendix:wasserstein}

\begin{figure}[h]
    \centering
    \includegraphics[width=\linewidth]{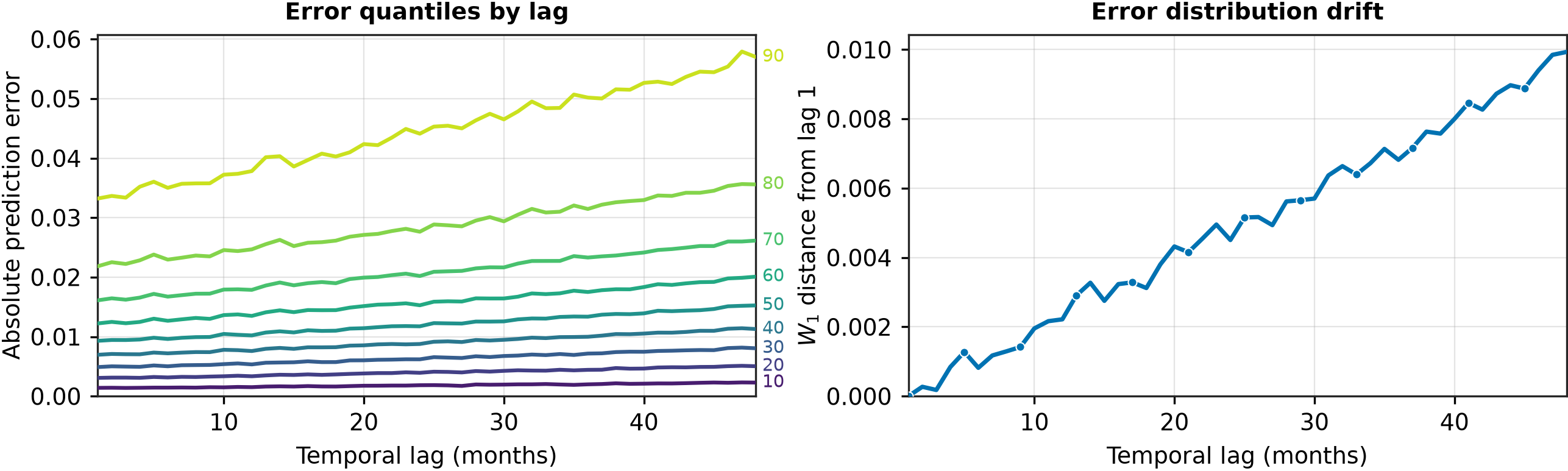}
    \caption{\textbf{Prediction error distributions smoothly degrade as evaluations grow stale.} \textit{(Left)} The 10th through 90th percentiles of absolute prediction error increase nearly linearly with monthly lag, with the largest increases occurring in the upper tail. \textit{(Right)} The 1-Wasserstein distance between each lag's error distribution and the 1-lag error distribution increases steadily with monthly lag.}
    \label{fig:wasserstein}
\end{figure}

We quantify the shift in predictive error distributions of 1-lag through 48-lag monthly predictors by plotting the 1-Wasserstein distance between each error distribution and the 1-lag error distribution (\cref{fig:wasserstein}, right). Distance increases steadily with monthly lag, by 0.001 over 48 months, which is $25\times$ larger than the median month-to-month 1-Wasserstein distance. Along with the continuity of the heatmap, this shows that the error distributions of lagged predictors shift smoothly with lag. This smooth distributional shift corresponds to increasing prediction error as evaluations grow stale. We demonstrate this in \cref{fig:wasserstein} (left), where the 10th through 90th percentiles of absolute error increase nearly linearly with lag (Spearman 0.986--0.997). The upper quantiles show the largest increases, with the 90th percentile prediction error increasing by 0.025. Thus, stale evaluations increase not only average error but also substantially increase tail risk. 

\newpage

\section{Annual and Monthly Forecasting Results}
\label{appendix:all-forecasters}

\begin{figure}[h]
    \centering
    \includegraphics[width=\linewidth]{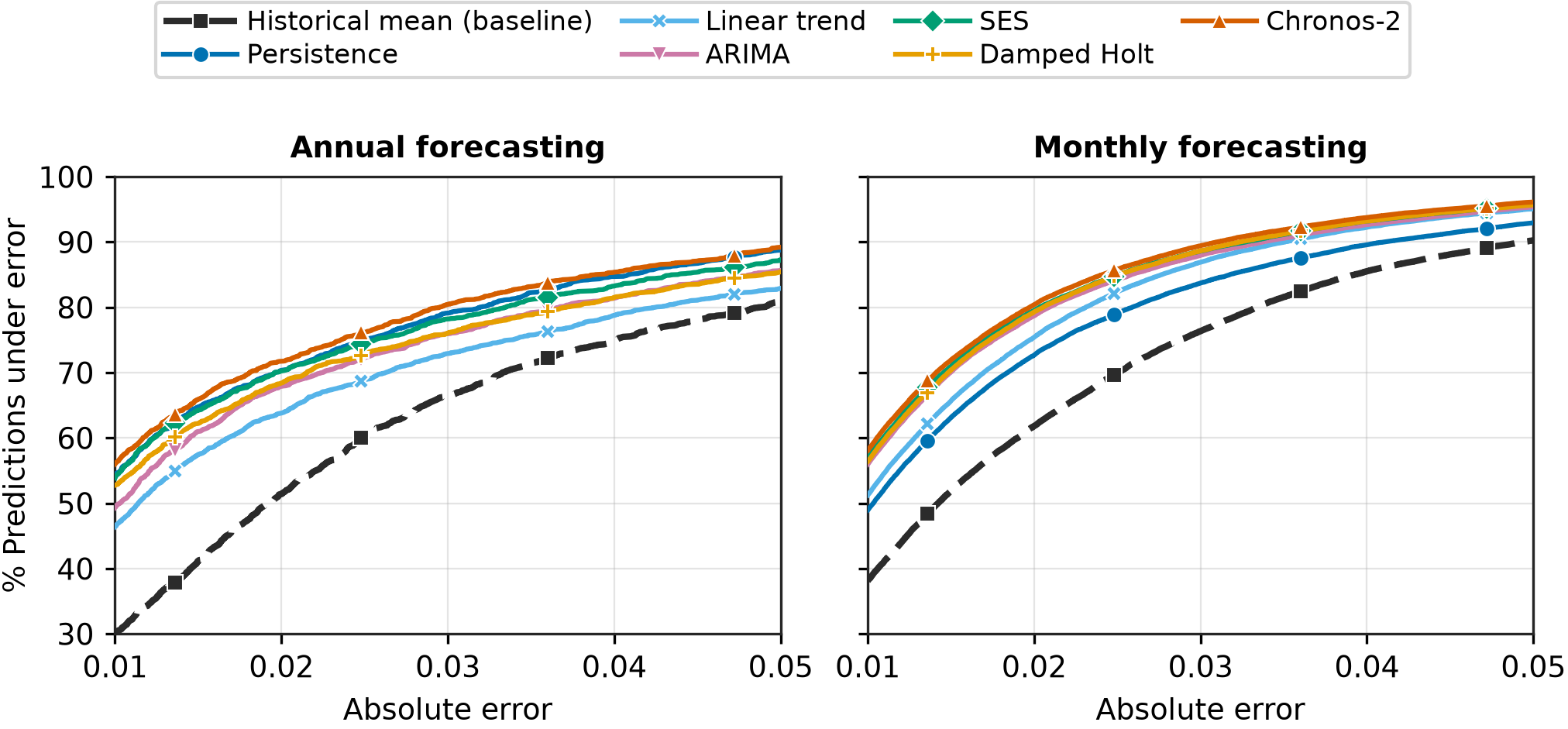}
    \caption{\textbf{Time-aware forecasters outperform historical mean pooling, with explicit trajectory modeling providing greater benefits at monthly resolution.} We plot prediction error CDFs for one-step-ahead forecasts at \textit{(left)} annual and \textit{(right)} monthly resolutions. At annual resolution, persistence performs nearly as well as Chronos-2, while at monthly resolution, explicit time series forecasters like Chronos-2 and SES improve substantially over persistence and perform similarly.}
    \label{fig:all-forecast}
\end{figure}

\begin{table}[h]
\centering
\caption{\textbf{Chronos-2 achieves the lowest MAE at annual and monthly resolutions, although several time-aware forecasters have comparable performance.} Mean absolute error (MAE) is for one-step-ahead forecasts of model performance. Lower is better.}
\label{tab:forecasting-results}
\setlength{\tabcolsep}{5pt}
\begin{tabular}{lcc}
\toprule
\textbf{Method} & \textbf{Annual MAE} & \textbf{Monthly MAE} \\
\midrule
Historical mean (baseline)           & 0.03083 & 0.02163 \\
Linear trend                & 0.02755 & 0.01527 \\
Damped Holt                 & 0.02360 & 0.01388 \\
ARIMA                       & 0.02266 & 0.01425 \\
SES                         & 0.02085 & \underline{0.01373} \\
Persistence (previous obs.) & \underline{0.02001} & 0.01735 \\
Chronos-2                   & \textbf{0.01938} & \textbf{0.01336} \\
\bottomrule
\end{tabular}
\end{table}

\newpage

\section{Forecasting and Robustness in Modern Coding Benchmarks}
\label{appendix:forecast-lcb}

We apply our previous monthly forecasting setup to LiveCodeBench\footnote{As before, we restrict our analysis to target months with a contiguous history of at least 12 months.} in \cref{tab:lcb_forecasting}. Unlike our broader monthly analysis with 12 datasets, where explicit trajectory modeling improves upon persistence, the most recent observation here already contains most of the available predictive signal.

\begin{table}[h]
    \centering
    \caption{\textbf{Persistence achieves the lowest MAE at monthly resolution on LiveCodeBench, although several time-aware forecasters have comparable performance.} Mean absolute error (MAE) is for one-step-ahead forecasts of model performance. Lower is better.}
    \label{tab:lcb_forecasting}
    \begin{tabular}{lr}
        \toprule
        \textbf{Method} & \textbf{Monthly MAE} \\
        \midrule
        Historical mean (baseline) & 0.15486 \\
        Linear trend & \underline{0.08539} \\
        Damped Holt & 0.08717 \\
        ARIMA & 0.10298 \\
        SES & 0.09071 \\
        Persistence (previous obs.) & \textbf{0.08506} \\
        Chronos-2 & 0.08837 \\
        \bottomrule
    \end{tabular}
\end{table}

Because LiveCodeBench contains related snapshots and reasoning variants of some models, has relatively few forecast target months, and includes problems that predate model training cutoffs, we test whether the persistence advantage is sensitive to these features. In addition to our standard historical mean baseline, we include a historical mean weighted by monthly problem counts. In \cref{tab:lcb_robustness}, we test whether the result depends on related model variants by equally weighting base-model families and leaving one family out, or on particular target months by leaving one month out. We also perform two cutoff-controlled analyses: one uses only post-cutoff evaluation months for each model; a stricter one also requires that all historical months used for a model's prediction postdate its cutoff. In all cases, persistence substantially outperforms both historical mean baselines.

\begin{table}[h]
    \centering
    \caption{\textbf{The advantage of persistence over the historical mean on LiveCodeBench is not sensitive to model-family composition, target-month choice, or training-cutoff restrictions.} Each sensitivity analysis report the reduction in persistence MAE relative to the usual historical mean and to a historical mean weighted by monthly problem counts.}
    \label{tab:lcb_robustness}
    \begin{tabular}{lrr}
        \toprule
        \textbf{Sensitivity analysis} & \textbf{Reduction vs. historical mean} & \textbf{Reduction vs. weighted mean} \\
        \midrule
        None & 45.1\% & 42.5\% \\
        Equal-weight families & 47.1\% & 44.7\% \\
        Leave one family out & 45.6--48.2\% & 43.1--45.8\% \\
        Leave one month out & 41.5--49.3\% & 38.7--46.7\% \\
        Post-cutoff months & 49.4\% & 47.4\% \\
        Post-cutoff months \& history & 39.0\% & 32.9\% \\
        \bottomrule
    \end{tabular}
\end{table}

A separate concern is that unequal monthly sample sizes could make some monthly scores noisier than others. We thus repeat our analysis after downsampling each month to 32 problems, excluding the unusually sparse final month (April 2025 has 12 problems). Across 1,000 resamples, 12-month-lag MAE exceeds one-month-lag MAE by 0.08 on average, and persistence reduces MAE relative to the historical mean by 36.6\% on average. Thus, neither the temporal locality result from \cref{appendix:lcb} nor the persistence advantage is explained by unequal monthly cohort sizes.

\newpage

\section{Model Ranking Inversions}
\label{appendix:inversions}

\begin{figure}[h]
    \centering
    \includegraphics[width=\textwidth]{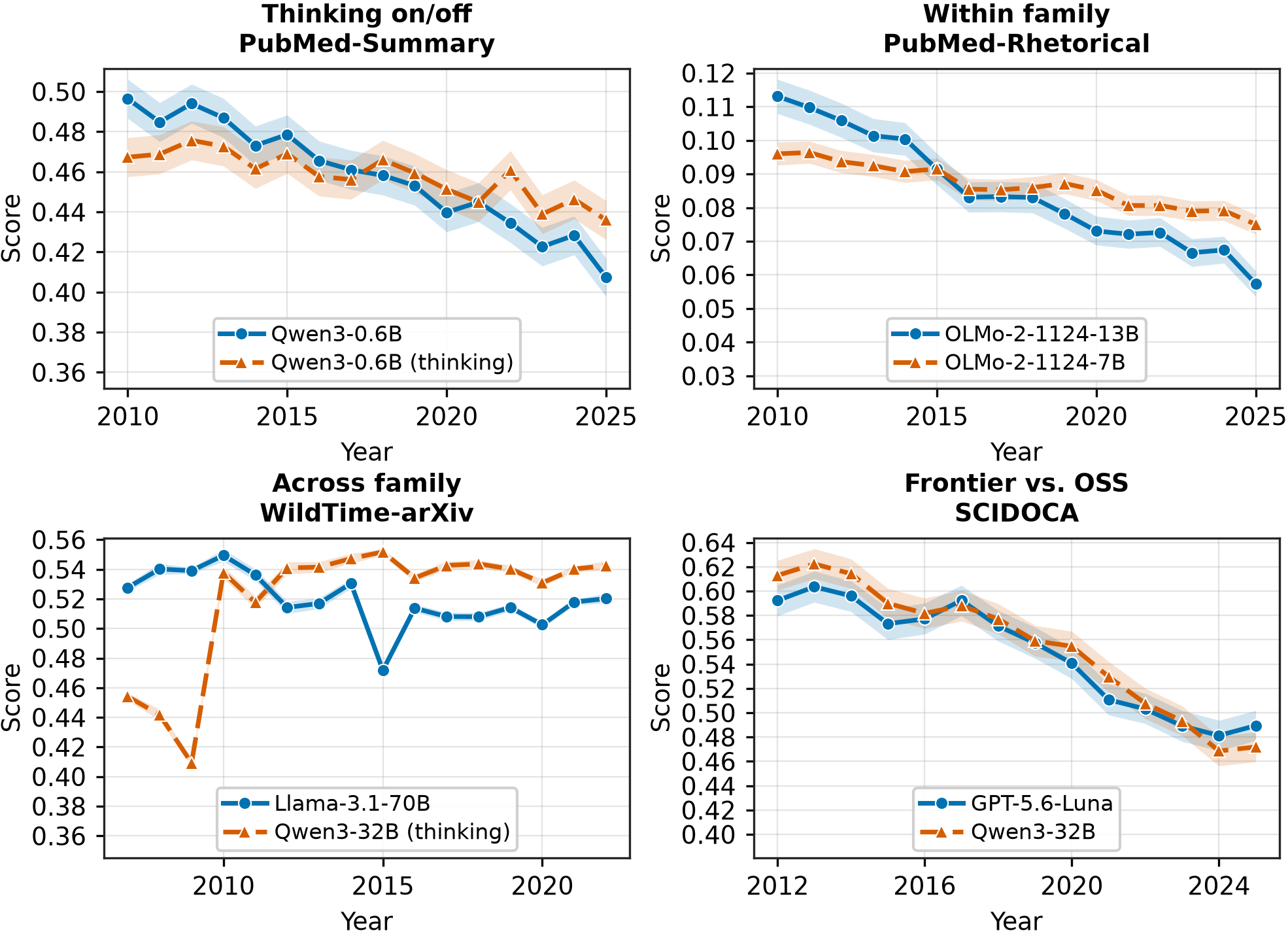}
    \caption{\textbf{Model rankings can reverse over time, causing temporally pooled scores to preserve outdated orderings.} Ranking inversions occur across years for thinking versus non-thinking variants, different model sizes within a family, similarly capable models across families, and frontier versus open-weight models. \textit{(Bottom left)} On WildTime-arXiv, the historical mean favors Llama-3.1-70B-Instruct even after Qwen3-32B-thinking becomes the better model in most later years.}
    \label{fig:inversions}
\end{figure}

WildTime-arXiv \citep{yao2022wild} provides a concrete example of historical pooling preserving an outdated model ranking (\cref{fig:inversions}, bottom-left). Ranking all 19 models by their historical mean selects Llama-3.1-70B-Instruct in all 11 target years, although Qwen3-32B-thinking is the best model in 10 of them. The pooled average remains dominated by evaluations predating the Qwen-Llama crossover, so the historical mean continues to reflect the older ranking. This is the most severe model selection failure due to staleness in our analysis.

\newpage

\section{Forecast Error and Model Selection}
\label{appendix:forecast-vs-select}

\begin{figure}[h]
    \centering
    \includegraphics[width=\linewidth]{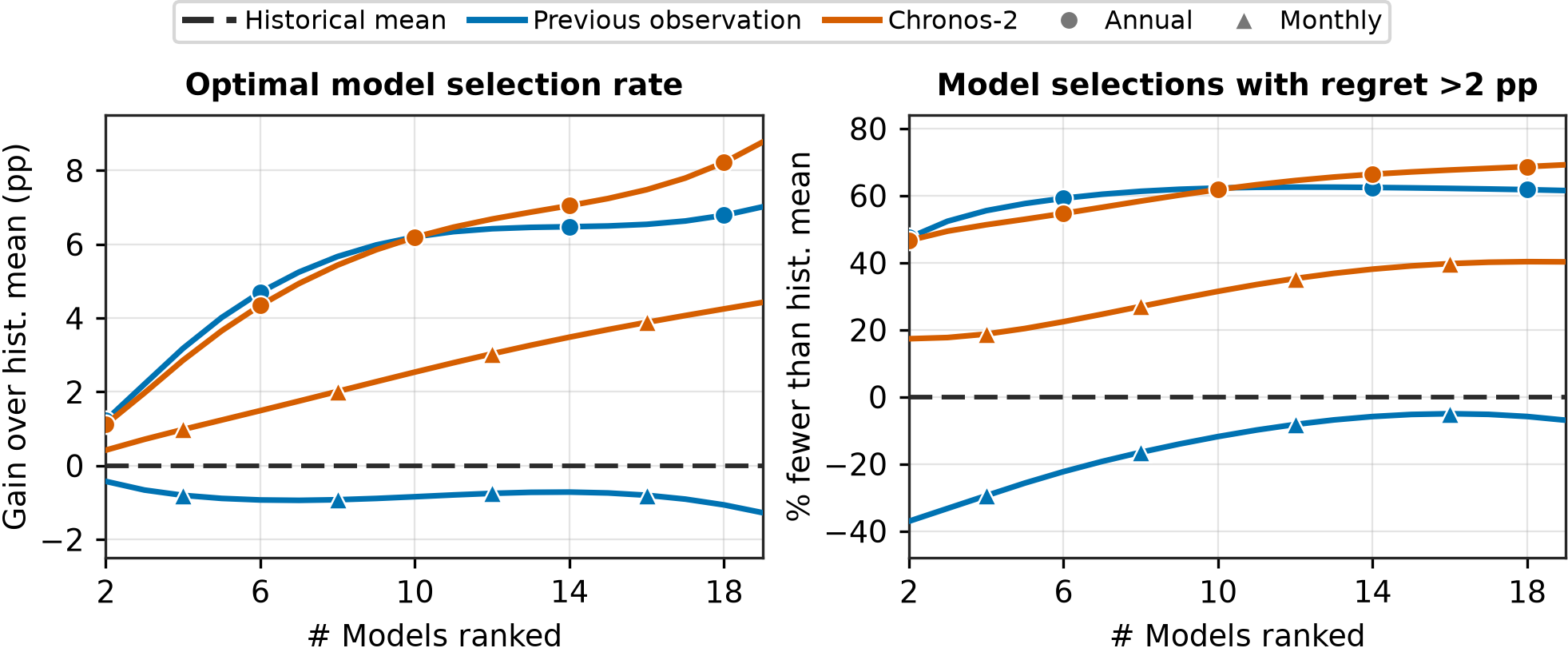}
    \caption{\textbf{Chronos-2 improves model selection at both annual and monthly resolutions, while persistence improves only annual selection.} As the number of candidate models increases, we plot \textit{(left)} absolute gains in optimal selection rate over the historical mean and \textit{(right)} reductions in model selections with regret greater than two percentage points relative to the historical mean. Circles are annual results, triangles are monthly results, and the dashed line is the historical mean baseline.}
    \label{fig:full-selection}
\end{figure}

At annual resolution, persistence and Chronos-2 both improve optimal model selection rates and reduce high-regret choices relative to the historical mean (\cref{fig:full-selection}). However, at monthly resolution, persistence is worse than the historical mean on both metrics despite reducing forecast MAE, whereas Chronos-2 improves both metrics relative to the historical mean. We diagnose this discrepancy by decomposing forecast error into a shared component that cannot affect model rankings and a residual component that can affect rankings.

For a dataset $b$, model $m$, and target period $t$, let
\begin{equation}
e_{b,m,t} = \hat{y}_{b,m,t} - y_{b,m,t}
\end{equation}
denote the forecast error. We decompose this error into the mean error across all $M=19$ models on the same dataset and target period, and a model-specific residual component:
\begin{equation}
e_{b,m,t}
=
\bar e_{b,t}+\tilde e_{b,m,t},
\qquad
\bar e_{b,t}
=
\frac{1}{M}\sum_{m=1}^{M}e_{b,m,t},
\qquad
\tilde e_{b,m,t}
=
e_{b,m,t}-\bar e_{b,t}.
\end{equation}
Because $\bar{e}_{b,t}$ (the shared error) is identical for every model, removing it cannot change their predicted ordering, including within any shortlist. However, it can affect forecast MAE. This decomposition helps explain the different relationships between forecast error and model selection at annual and monthly resolutions. Annually, persistence's MAE reduction relative to the historical mean falls from 35.1\% to 25.5\% after removing shared error, while Chronos-2's falls from 37.1\% to 23.0\%. At monthly resolution, persistence's MAE reduction relative to the historical mean falls from 19.8\% to only 3.3\%, while Chronos-2 retains a substantial MAE reduction, falling from 38.2\% to 23.3\%. Thus, much of persistence's monthly reduction in forecast error is shared across models and irrelevant to rankings. This shows why lower forecast MAE does not always translate into a higher optimal selection rate.

\newpage

\section{Temporal Locality in Multimodal Evaluation}
\label{appendix:vlm-graphs}

Our experiments thus far are specific to LLMs and text-only benchmarks. We now ask whether the same structure extends to multimodal evaluation, where distribution shift can arise from changes in both visual content and image-text relationships. We evaluate Gemma-3-12B-IT and Gemma-3-27B-IT on Fakeddit \citep{nakamura2019r}, a Reddit task where the goal is to classify pairs of images and post titles as true or fake. We use 10,000 balanced examples per year from 2012--2019.

\begin{figure}[h]
    \centering
    \includegraphics[width=\linewidth]{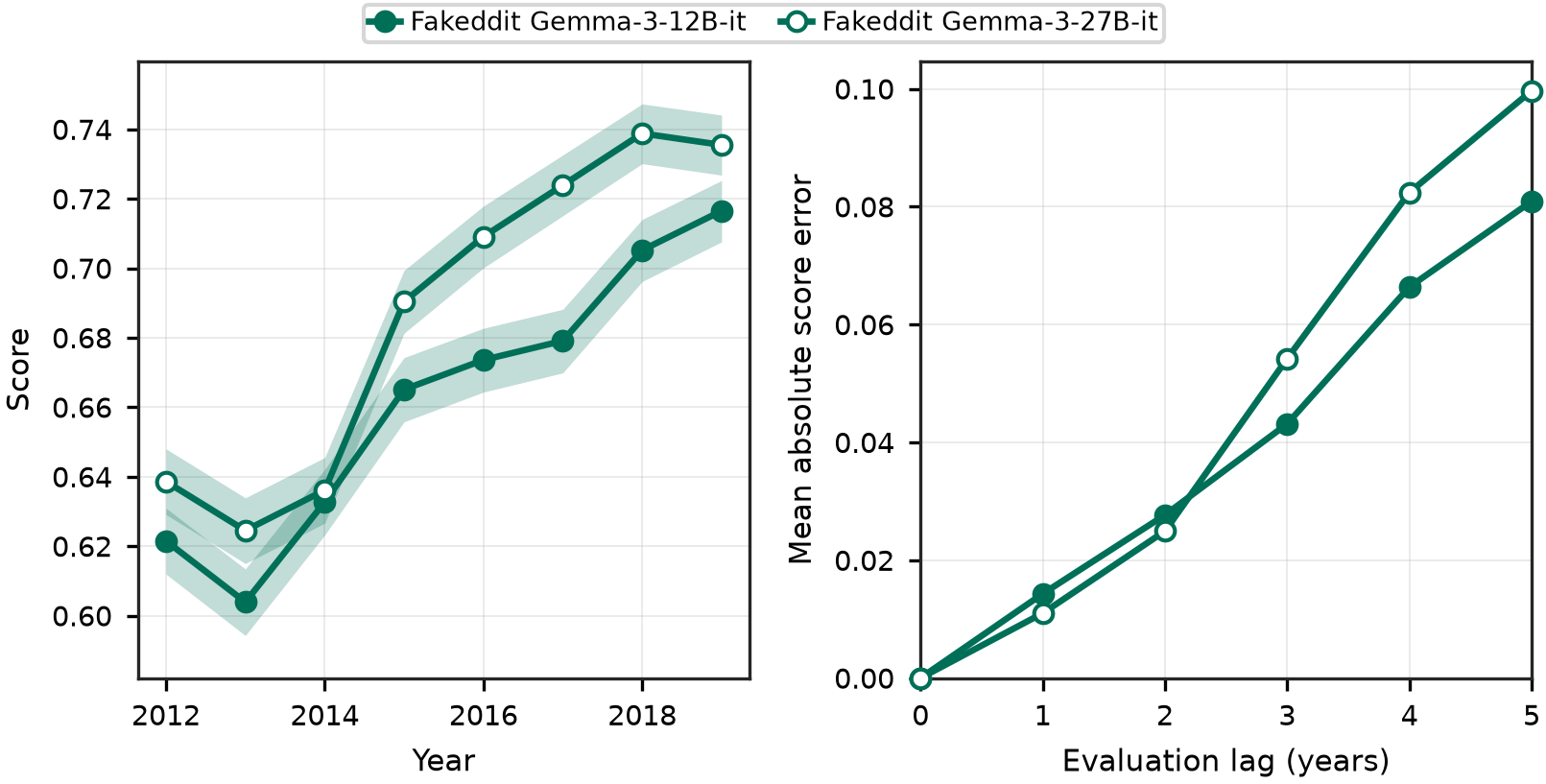}
    \caption{\textbf{Recent evaluations better predict future multimodal performance.} We evaluate Gemma-3-12B-IT and Gemma-3-27B-IT on Fakeddit, grouping performance by year. \textit{(Left)} Performance substantially increases over time for both models. \textit{(Right)} Prediction error increases with evaluation age, so recent evaluations are the best predictors of future performance.}
    \label{fig:vlm}
\end{figure}

We again find that recent evaluations are more predictive of future performance than stale ones. At annual resolution, predictive error increases consistently with evaluation age for both models (\cref{fig:vlm}). Because class balance is fixed across time, this pattern reflects changes within the underlying data distribution, such as evolving topics, visual styles, and image-text relationships. Further, all examples predate Gemma 3's August 2024 training cutoff, ruling out post-cutoff evaluation as a cause. Thus, temporal locality extends to multimodal performance, suggesting that it is a general property of evolving evaluation distributions rather than text alone.

\newpage

\section{Data Recency in Pretraining}
\label{appendix:training-data}

\begin{figure}[h]
    \centering
    \includegraphics[width=\linewidth]{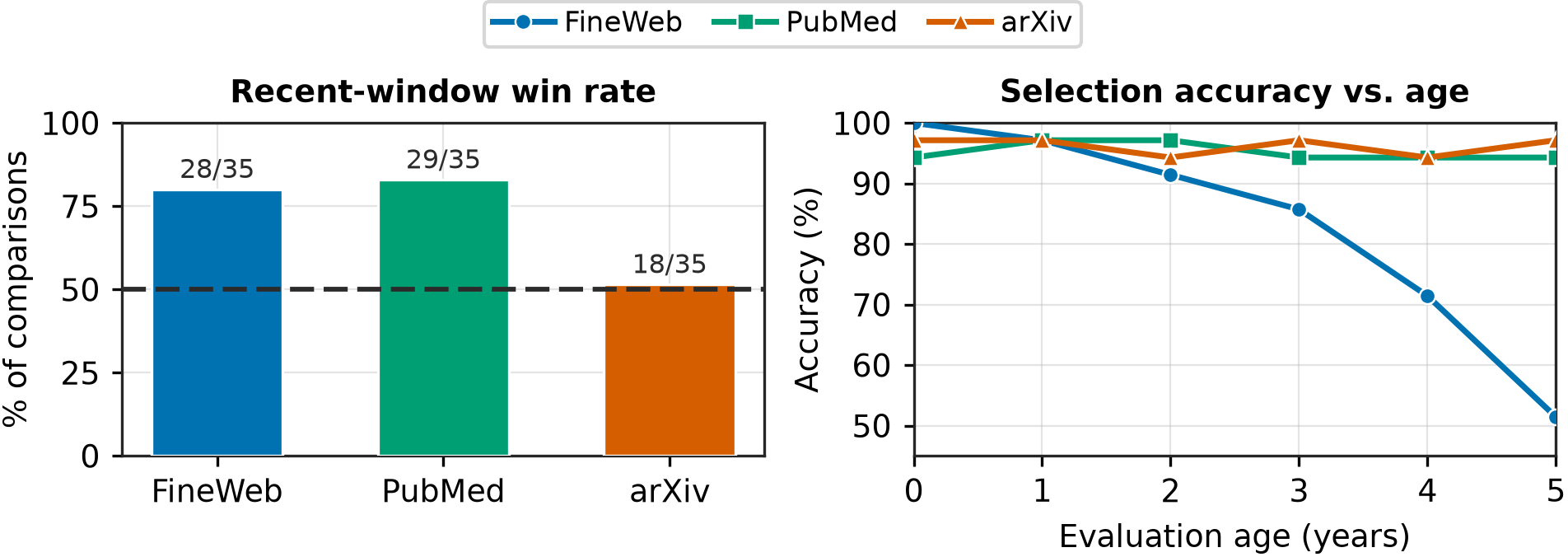}
    \caption{\textbf{The benefit of data recency in pretraining is domain-dependent.} \textit{(Left)} Recent-window pretraining often yields lower next-year perplexity than cumulative pretraining on FineWeb and PubMed, while arXiv shows no preference. \textit{(Right)} The most recent yearly target-domain evaluation accurately selects between the two mixtures across all three datasets. Selection accuracy sharply degrades with evaluation age on FineWeb but remains stable on PubMed and arXiv.}
    \label{fig:training-data-selection}
\end{figure}

\newpage

\section{Data Recency in Supervised Learning}
\label{appendix:active-learning}

\begin{figure}[h]
    \centering
    \includegraphics[width=\linewidth]{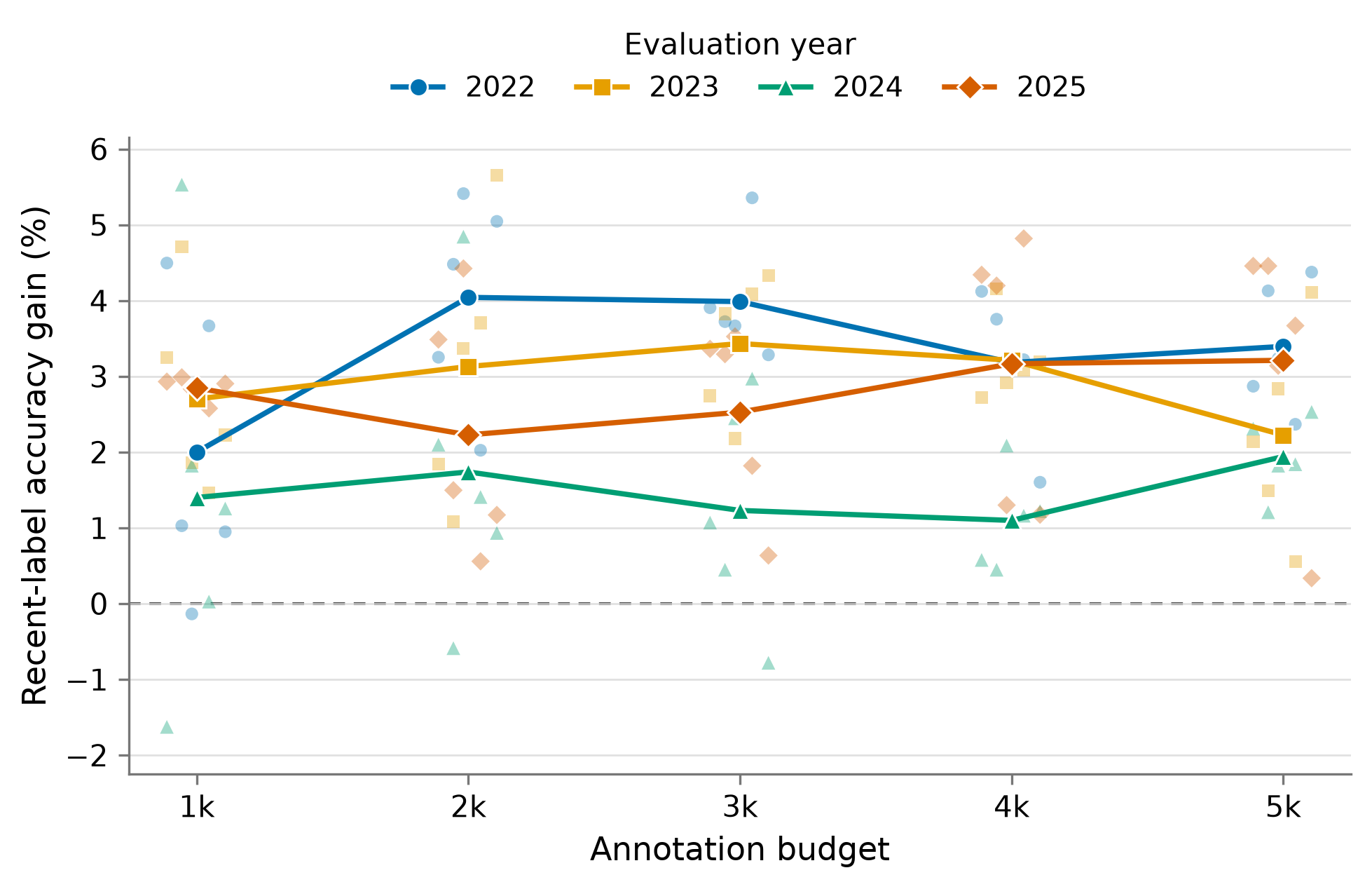}
    \caption{\textbf{Recent supervision improves next-year performance under a fixed annotation budget.}
    For each evaluation year from 2022 to 2025, we finetune Llama-3.1-8B-Instruct on NVD-RCM data uniformly sampled from either all preceding years (All) or the most recent year (Recent). Points show paired Recent-All next-year accuracy differences across five seeds, and lines show means. Recent sampling improves mean accuracy in all 20 year-budget settings, with an average gain of 2.63\%.}
    \label{fig:nvd_recent_labels}
\end{figure}

\end{document}